\documentclass{article} % For LaTeX2e
\usepackage{times}
\usepackage[accepted]{icml2024}

\usepackage{amsmath,amsfonts,bm}

\def\eqref#1{equation~\ref{#1}}
\def\1{\bm{1}}

\DeclareMathAlphabet{\mathsfit}{\encodingdefault}{\sfdefault}{m}{sl}
\SetMathAlphabet{\mathsfit}{bold}{\encodingdefault}{\sfdefault}{bx}{n}

\usepackage{microtype}
\usepackage{graphicx}
\usepackage{subfigure}
\usepackage{booktabs} % for professional tables
\usepackage{adjustbox}

\usepackage{algorithm, algorithmic}

\usepackage[utf8]{inputenc} % allow utf-8 input
\usepackage[T1]{fontenc}    % use 8-bit T1 fonts
\usepackage[breaklinks,colorlinks]{hyperref}

\usepackage{graphicx}
\usepackage{amsmath}
\usepackage{amssymb}
\usepackage{booktabs}
\usepackage[utf8]{inputenc} 

\usepackage{adjustbox}

\usepackage[nameinlink]{cleveref}
\Crefname{section}{Section}{Sections}
\Crefname{table}{Table}{Tables}
\Crefname{figure}{Figues}{Figues}
\usepackage{pifont}

\usepackage{url}

\usepackage{tikz}
\usepackage{comment}
\usepackage{amsfonts}       % blackboard math symbols
\usepackage{color}

\usepackage{amsmath}
\usepackage{amssymb}
\usepackage{bbm, dsfont}
\usepackage{mathtools}
\usepackage{amsthm}
\usepackage[normalem]{ulem}
\usepackage{xcolor,colortbl}
\usepackage{multirow}
\usepackage{cutwin}
\usepackage{arydshln}
\usepackage{caption}
\usepackage{enumitem}
\usepackage{wrapfig}
\usepackage{microtype}
\usepackage{graphicx}
\usepackage{bm}
\usepackage{eqparbox}
\usepackage{makecell}
\usepackage{threeparttable}

\renewcommand{\algorithmiccomment}[1]{\hfill\eqparbox{COMMENT}{\footnotesize $\rhd$ #1}}
\newcommand{\eat}[1]{}

\definecolor{Red}{rgb}{0.6,0,0}
\definecolor{Blue}{rgb}{0,0,0.8}
\definecolor{Green}{rgb}{0,0.4,0.7}
\definecolor{airforceblue}{rgb}{0.36, 0.54, 0.66}
\definecolor{ao(english)}{rgb}{0.0, 0.5, 0.0}
\definecolor{azure(colorwheel)}{rgb}{0.0, 0.5, 1.0}
\definecolor{crimson}{rgb}{0.86, 0.08, 0.24}
\definecolor{darkcerulean}{rgb}{0.03, 0.27, 0.49}
\definecolor{cobalt}{rgb}{0.0, 0.28, 0.67}
\definecolor{rosegold}{rgb}{0.72, 0.43, 0.47}
\definecolor{orange-red}{rgb}{1.0, 0.27, 0.0}
\definecolor{mountainmeadow}{rgb}{0.19, 0.73, 0.56}
\definecolor{malachite}{rgb}{0.04, 0.85, 0.32}
\definecolor{darkblue}{rgb}{0.0, 0.0, 0.55}
\definecolor{customblue}{rgb}{0.2, 0.35, 0.8}
\definecolor{gg}{gray}{0.92}
\newcolumntype{a}{>{\columncolor{gg}}c}

\definecolor{gg}{gray}{0.95}
\definecolor{ggg}{gray}{0.55}
\hypersetup{colorlinks=true}
\hypersetup{linktoc=all}
\hypersetup{citecolor=mountainmeadow}
\hypersetup{linkcolor=crimson}
\hypersetup{urlcolor=darkblue}
\usepackage[all]{hypcap}

\creflabelformat{equation}{#2\textup{#1}#3}  

\usepackage{svg}

\Crefname{section}{Sec.}{Secs.}
\Crefname{equation}{Eq.}{Eqs.}
\Crefname{table}{Tab.}{Tabs.}
\Crefname{figure}{Fig.}{Figs.}

\newcommand{\highlight}[1]{{\color{crimson}{#1}}}
\newcommand{\colorhi}[1]{{\color{azure(colorwheel)}{#1}}} 
\newcommand{\evblue}[1]{{\color{azure(colorwheel)}{#1}}}

\icmltitlerunning{EVEREST: Efficient Masked Video Autoencoder by Removing Redundant Spatiotemporal Tokens}

\begin{document}
\twocolumn[
\icmltitle{\includegraphics[width=0.04\linewidth]{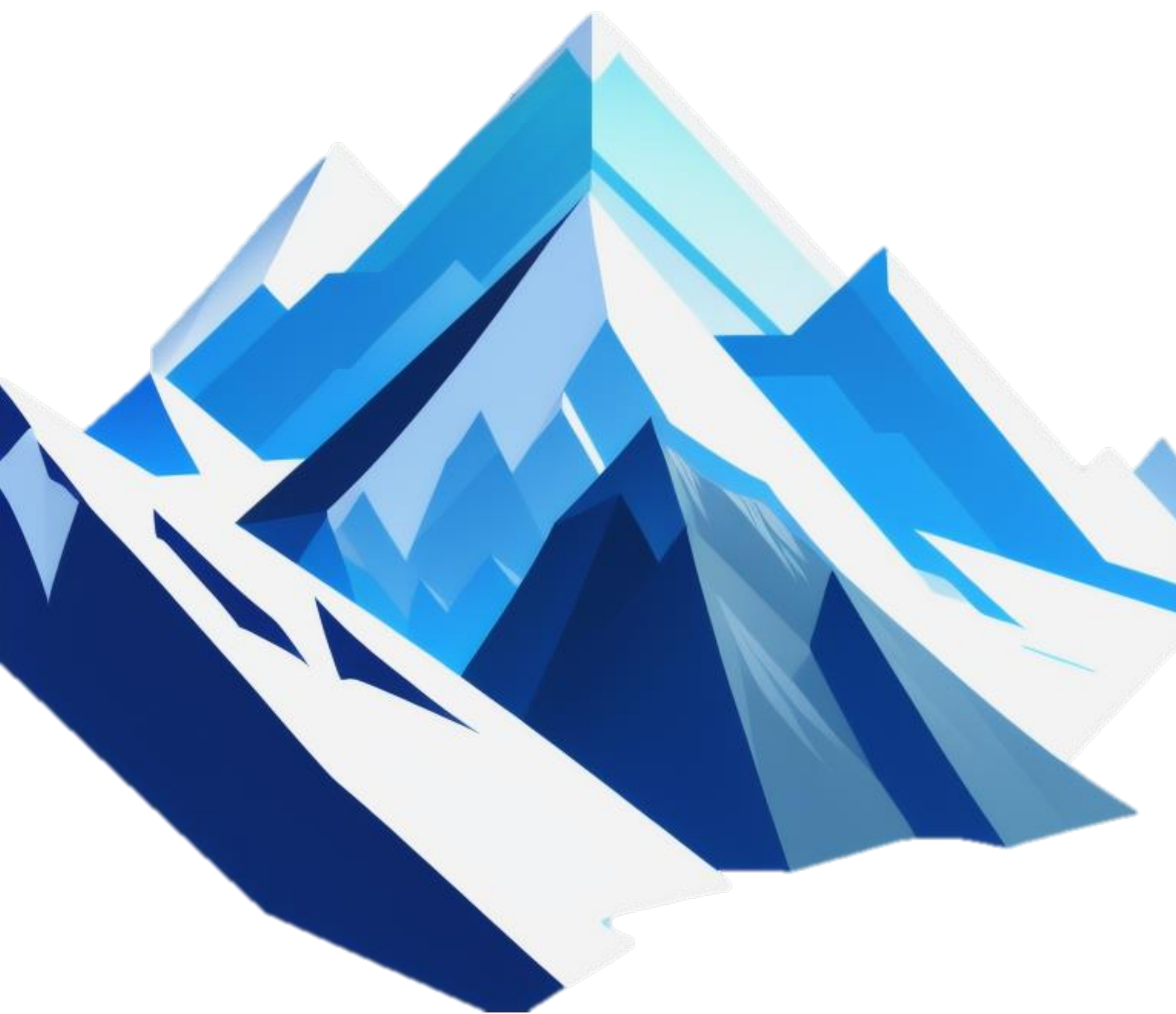}~EVEREST: Efficient Masked Video Autoencoder by\\ Removing Redundant Spatiotemporal Tokens}

\icmlsetsymbol{equal}{*}

\begin{icmlauthorlist}
\icmlauthor{Sunil Hwang}{equal,kma}
\icmlauthor{Jaehong Yoon}{equal,unc}
\icmlauthor{Youngwan Lee}{equal,kaist,etri}
\icmlauthor{Sung Ju Hwang}{kaist,deepauto}
% Sunil Hwang$^{1*}$ \;\; Jaehong Yoon$^{2*}$ \;\; Youngwan Lee$^{1*}$\thanks{Equal contribution} \;\; Sung Ju Hwang$^{1,3}$ \\
%   KAIST$^{1}$,\; UNC Chapel Hill$^{2}$,\;DeepAuto$^{3}$\; \\
%   \texttt{sunilhoho93@gmail.com, jhyoon@cs.unc.edu, \{ywlee88, sjhwang82\}@kaist.ac.kr}
\end{icmlauthorlist}

\icmlaffiliation{kma}{Korea Military Academy}
\icmlaffiliation{kaist}{KAIST}
\icmlaffiliation{unc}{UNC Chapel Hill}
\icmlaffiliation{deepauto}{DeepAuto}
\icmlaffiliation{etri}{ETRI}

\icmlcorrespondingauthor{Sunil Hwang}{sunil.hwang93@kma.ac.kr}
\icmlcorrespondingauthor{Sung Ju Hwang}{sjhwang82@kaist.ac.kr}

\icmlkeywords{Machine Learning, ICML}
\vskip 0.3in

]
% \printAffiliationsAndNotice{}
\printAffiliationsAndNotice{\icmlEqualContribution} % otherwise use the standard text.

\begin{abstract}
Masked Video Autoencoder (MVA) approaches have demonstrated their potential by significantly outperforming previous video representation learning methods. However, they waste an excessive amount of computations and memory in predicting uninformative tokens/frames due to random masking strategies. (e.g., over 16 nodes with 128 NVIDIA A100 GPUs~\citep{feichtenhofer2022masked}). To resolve this issue, we exploit the unequal information density among the patches in videos and propose \textit{\textbf{EVEREST}}, a \textit{surprisingly efficient} MVA approach for video representation learning that finds tokens containing rich motion features and discards uninformative ones during both pre-training and fine-tuning. We further present an information-intensive frame selection strategy that allows the model to focus on informative and causal frames with minimal redundancy. Our method significantly reduces the computation and memory requirements of MVA, enabling the pre-training and fine-tuning on \textbf{a single machine with 8 GPUs} while achieving comparable performance to computation- and {memory-heavy baselines} on multiple benchmarks and the uncurated Ego4D dataset. We hope that our work contributes to reducing the barrier to further research on video understanding. Codes are available at: \href{https://github.com/sunilhoho/EVEREST}{\textcolor{magenta}{https://github.com/sunilhoho/EVEREST}}.
\end{abstract}

\begin{figure*}[h]
    \centering
    \begin{minipage}{0.43\textwidth}
    \centering
    \setlength{\tabcolsep}{-1pt}{%
    \includegraphics[width=1.0\linewidth, trim={0.0cm 0.0cm 0.0cm 0.0cm},clip]{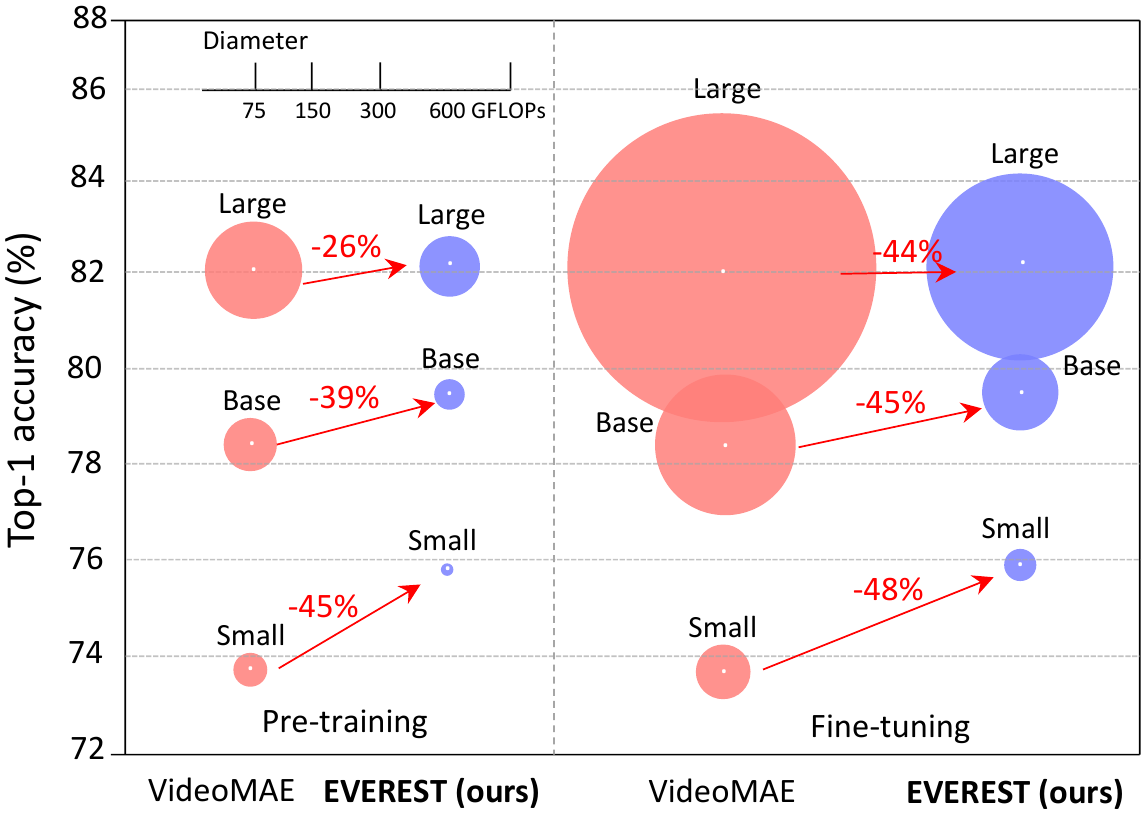}%\label{fig:flop_plot}
    }
    \textbf{(a) Computation~(GFLOPs) Comparison}
    \end{minipage}
    \hspace{0.5cm}
    \begin{minipage}{0.41\textwidth}
    \centering
    \setlength{\tabcolsep}{-1pt}{%
    \vspace{0.1in}
    \includegraphics[width=1.0\linewidth, trim={0.0cm 0.0cm 0.0cm 0.0cm},clip]{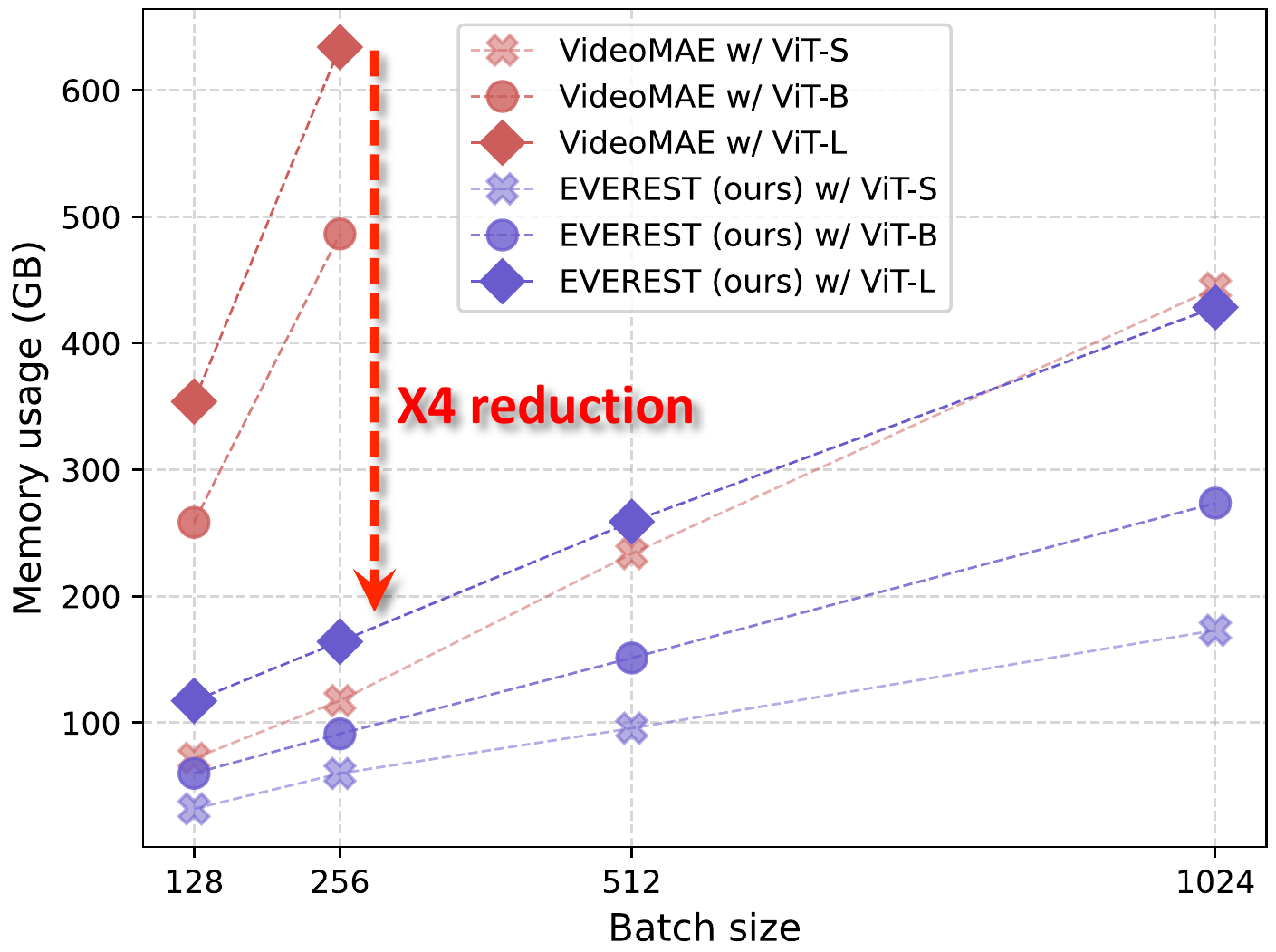}%\label{fig:pt_mem}
    }
    \textbf{(b) Memory comparison}
    \end{minipage}
    \caption{\textbf{Efficiency of our EVEREST against VideoMAE on K400 dataset.}     \textbf{(a) GFLOPs} for pre-training and fine-tuning. The bubble size is proportional to the GFLOPs of the model. \textbf{(b) Memory consumption} using one node equipped with 8$\times$ A100~(80GB). VideoMAE with ViT-B and -L fails to deploy the model due to \textit{\textbf{out-of-memory}} if the batch sizes are 512 or larger. 
    For a ViT-L backbone with a batch size of 256, our method achieves about $4 \times$ less memory consumption than VideoMAE.    
    Please see ~\Cref{tab:vs_videomae,tab:mem-allocation2} for detailed results.
    }
    % \vspace{-0.15in}
    \label{fig:resource}
\end{figure*}

\section{Introduction}\label{introduction}
Massive video data floods the web and media daily with the rapid growth of portable devices equipped with cameras, such as AR glasses, smartphones, UAVs, and robots. However, direct utilization of user-generated video data to solve target task problems is nontrivial, as annotating them is time-consuming and expensive. 
One potential approach to tackle this problem is to learn generic representations from unlabeled video streams that can transfer to diverse downstream visual tasks. Video Representation Learning~(VRL)~\citep{fernando2017self, piergiovanni2020evolving, qian2021spatiotemporal, pan2021videomoco} methods allow learning spatial and temporal features from input video frames in a self-supervised manner without any human annotations. 
A caveat to such pre-training for video tasks is that, unlike image-based tasks with static information of objects in instances, video-based tasks involve temporal causality; that is, successive frames are strongly correlated in their semantics.

Recently, Masked Video Autoencoder (MVA)~\citep{tong2022videomae,feichtenhofer2022masked}, which learns to reconstruct randomly masked spatiotemporal regions in video clips, has shown impressive performance on various video-based problems. Yet, critical challenges remain in efficiently exploiting the spatiotemporal information in real-world videos:
(1) Tokens (a pair of two temporally successive patches in the same space) in videos are not equally valuable to reconstruct, as the amount of their information depends not only on spatial importance but also on temporal redundancy. (2) Learning representations from videos is infeasible without a huge computing budget. MVA approaches~\citep{feichtenhofer2022masked,wang2023masked,wang2023mvd} that reconstruct the \textit{whole} video require excessively large amounts of computations, making it impractical without access to a substantial GPU cluster. For example, VideoMAE~\citep{tong2022videomae} takes about 27 hours to pre-train for 800 epochs with ViT-B using \colorhi{\textbf{64 NVIDIA V100~(32GB) GPUs}}, and ST-MAE~\citep{feichtenhofer2022masked} takes about 35.8 hours to pre-train for 800 epochs with ViT-L using \colorhi{\textbf{128 NVIDIA A100~(80GB) GPUs}}.

To overcome such limited feasibility of video representation learning, we propose an \textit{\textbf{E}fficient Masked \textbf{V}ideo Auto\textbf{E}ncoder which removes \textbf{RE}dundant \textbf{S}patiotemporal \textbf{T}okens (\textbf{EVEREST})}, that is a highly efficient training algorithm with token selection to alleviate the extremely high resource requirements for VRL. Unlike previous methods~\citep{tong2022videomae,feichtenhofer2022masked,wang2023videomae} reconstructing \emph{all} patches, we aim to select a subset of \emph{informative} visual tokens to learn based on the distance disparity across temporally adjacent tokens in the embedding space. 
Our \textit{redundancy-robust token selection} approach successfully detects meaningful changes in the state/object of incoming videos in an online manner,  discarding less meaningful tokens containing redundant information or meaningless backgrounds, without resorting to dense motion features from the incoming video, such as HOG or optical flows~\citep{sun2023mme,wang2023mvd}.

This allows us to significantly reduce computational cost and memory usage while maintaining the quality of a representation model by backpropagating to only a few selected tokens that retain rich spatiotemporal information with minimal redundancy. As shown in~\Cref{fig:resource}, our EVEREST \textbf{saves computation costs by} \textbf{$\bf 26\sim45$\%} in pre-training and \textbf{$\bf 44\sim48$\%} in fine-tuning across varying ViT scales while achieving competitive performance against strong VideoMAE baselines. We also achieve an impressive amount of memory reduction with all ViT backbones, which becomes more effective with larger models. For example, on a ViT-Large with a batch size of 256, our method achieves about \textbf{$\bf 4 \times$ smaller memory consumption} than VideoMAE and \textbf{enables single-node training with larger batch sizes} using a large backbone, whereas existing VRL methods require immense memory occupancy and fail to train in the same setup as they go \textit{out-of-memory}.  %We also aim to enhance the accessibility of those cutting-edge video models by employing practical resources on a single node and pursuing eco-friendly AI research.

In addition, most current VRL~\citep{arnab2021vivit,bertasius2021timesformer,feichtenhofer2022masked,tong2022videomae} methods uniformly load frames at regular time intervals from each mini-batch clip. However, this strategy does not consider a temporal imbalance in information and noise that matters to real-world uncurated videos. Our EVEREST further performs \textit{information-intensive frame selection}, which is carried out online through \textit{probabilistic sampling} proportional to the ratio of redundancy-robust tokens in each frame and does not need any additional computational or parametric learning steps. Consequently, we can capture abundant spatiotemporal knowledge from highly sparse yet informative spatiotemporal regions in videos.

We extensively validate our proposed method on multiple benchmark datasets, including UCF101, HMDB51, K400, Something-Something V2, and Ego4D, and our EVEREST shows remarkable efficiency in terms of memory occupancy, computational cost, and training time compared to strong counterparts, achieving competitive performance. 

Our contributions are as follows:
\begin{itemize}[leftmargin=0.25in]
\item We propose \textit{redundancy-robust token selection}, an \textit{efficient} VRL method that promptly selects the most informative tokens based on the states' change and discards redundant ones in an online manner, avoiding wasteful training on uninformative regions of videos.

\item We further propose \textit{information-intensive frame selection}, a strategy to select informative video frames from incoming videos, which allows the model to efficiently learn robust and diverse temporal representations in real-world uncurated videos. 

% This is particularly helpful for training real-world uncurated videos~(e.g., Ego4D).

\item Our EVEREST has great potential to lower the barrier for video-related research that requires enormous computing power and cost, as it shows comparable performance to existing methods while significantly reducing the computations, memory, and training time.

\end{itemize}
\section{Related Work}\label{related_work}
\paragraph{Masked video autoencoder}
Inspired by self-supervised learning with Masked Image Modeling~\citep{he2022masked,xie2021simmim,kakogeorgiou2022hide}, several recent works on video representation learning~\citep{wang2022bevt,sun2023mme,wang2023videomae,wang2023mvd} suggest spatiotemporal masking strategies given video streams.
To capture spatial representation and temporal dynamics for unsupervised video streams, ST-MAE~\citep{feichtenhofer2022masked} and VideoMAE~\citep{tong2022videomae} extend a masked image autoencoder to mask partial regions in arriving video clips via random and space-only random sampling, respectively. They find that spatiotemporal inductive bias in video clips helps a decoder predict input pixels in masked regions, allowing a higher masking ratio ($\sim90\%$) than MIM {($\sim60\%$~\citep{xie2021simmim} and $\sim75\%$~\citep{he2022masked})} on image self-supervised learning. 
BEVT~\citep{wang2022bevt} proposes to train image- and video-level masked autoencoders jointly by sharing weights of the encoder, formulated with Video Swin~\citep{liu2022video}. They resort to random sampling given spatiotemporal inductive bias, which can be a good approximator with stochasticity during data-driven training. 
Nevertheless, selecting random tokens to reconstruct for Masked video autoencoder is inefficient since embedded tokens in video frames are not equally important, especially since the informativeness of each token is affected by the adjacent frames.

% \vspace{-0.1in}
\paragraph{Input selective training} 
As benchmark training datasets often have massive scales and contain a lot of redundant or less meaningful instances, various works~\citep{fayyaz2022ats,wu2019adaframe,yoon2022online} have discussed sampling important instances from the entire dataset or focusing on localized information in each frame~\citep{meng2022adavit,fayyaz2022ats,kakogeorgiou2022hide,yin2022vit} for efficient image recognition. However, they have no means to capture a temporal correlation across adjacent frames in video tasks. 
A few works recently suggest \textit{supervised input selection} techniques for video tokens or frames. \cite{wang2021efficient} introduce a lightweight scorer network to select the most informative temporal and spatial token in incoming videos for supervised video classification tasks. \cite{Park_2022_ECCV} suggest the greedy K-center search that iteratively finds the most distant patches in the geometrical space from video clips. %\cite{korbar2019scsampler} jointly train the clip classifier and clip-level saliency model for gathering the clips most useful to the clip classifier. 
\cite{gowda2021smart} train a single and global frame selector based on the ground truth for computing the importance score of single and paired frames. 
\cite{zhi2021mgsampler} performs a frame selection for unsupervised videos, but they have to extract whole frames from the training video dataset in advance to compute the change of cumulative distribution in video frames and features, consuming substantial pre-processing time and storage for saving the extracted frames (it takes {two days} to extract SSv2 into frames and occupies {433 GB}). Similarly, a few recently proposed MVA methods with adaptive token sampling require extracting dense motion information in advance or learnable parameters. MGM~\citep{fan2023motion}, MGMAE~\citep{huang2023mgmae} and MME~\citep{sun2023mme} generate motion-guided masking maps to reconstruct the informative tokens of the given videos, but they require motion vectors and optical flows, respectively. 
% AdaMAE~\citep{bandara2023adamae} trains the sampling layer, which predicts the probability of the masking for each token, but it requires additional training costs. 

However, {extracting all video frames and computing their motion information (e.g., HoG or optical flows) in advance is unrealistic} for online frame selection in videos. On the other hand, our EVEREST can perform rapidly without time- and memory-wasting steps before training.

% Recently proposed adaptive sampling-based MVM methods show better performance; however, they still require dense motion information or learnable parameters for adaptive sampling.  Unlike the above, our method does not require additional dense information and training costs while showing compatible performance.

% Unlike earlier methods \textbf{requiring labeled information in videos} for adaptive token selection, our method can rapidly select informative spatiotemporal tokens and frames without any supervision by leveraging embedded motion information. In the end, we drastically reduce the required computational cost and memory from both the pre-training and fine-tuning stages through information-intensive token selection. 

% AdaVit~\cite{meng2022adavit} introduces parametric decision networks for patch/head/block selection for training ViT. ATS~\cite{fayyaz2022ats} samples meaningful tokens based on the significant score, computed by multiplication of attentions and value matrices. AttMask~\cite{kakogeorgiou2022hide} and A-Vit~\cite{yin2022vit} adopt parameter-free token masking strategies utilizing the average of multi-head attentions. 
% \cite{wu2019adaframe} trains the memory-augmented LSTM for gathering informative frames of the given video and its ground truth. %The training step requires the ground truth of the given video to evaluate whether observing more frames is expected to produce more accurate predictions.
\begin{figure*}
% \vspace{-0.08in}
    % \small
    \centering
    \includegraphics[width=\linewidth]{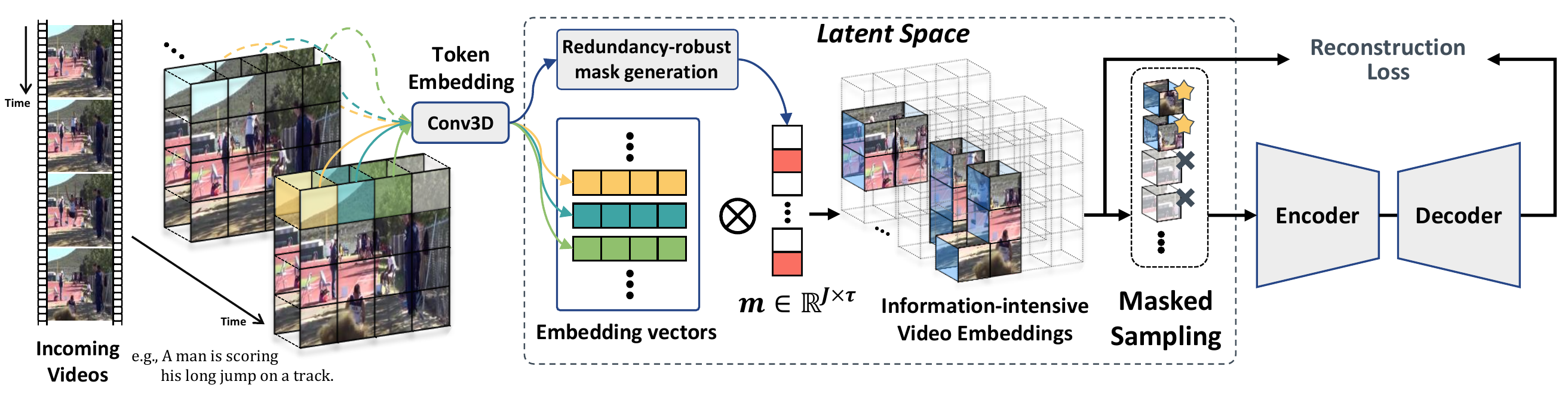}
    % \vspace{-0.15in}
     \caption{\textbf{Overview of EVEREST}. Our redundancy-robust mask generator selects tokens with a large disparity with the paired ones in the previous time dimension, indicating that they include rich motion features. Then, the model focuses on learning representation by reconstructing only sparsified videos containing abundant spatiotemporal information, which makes the VRL surprisingly efficient.
     }
    \label{fig:main-concept-figure}
% \vspace{-0.05in}
\end{figure*}

\section{Masked Video Autoencoders}
\subsection{Masked Video Autoencoder}\label{subsec:MVA}
% Learning to reconstruct intentionally corrupted data with masking is broadly utilized as means of representation learning in Natural Language Processing~\cite{devlin2018bert, song2019mass, guu2020retrieval, song2020mpnet} and has demonstrated its efficacy and power in broad research problems. In vision tasks,
Masked Image Modeling (MIM)~\cite{he2022masked, xie2021simmim} aims to learn representations of the input images by solving the regression problem in which a model predicts RGB pixel values in randomly masked patch regions of images. %The model divides an image into equally sized patches and then randomly chooses them to be masked based on a predetermined ratio. Given unmasked patches, the encoder transforms them into feature vectors, and a decoder aims to reconstruct the original input images by predicting the RGB values of the missing image patches.
The recent success of MIM has led to breakthroughs in effectively capturing spatiotemporal information from incoming video streams, which we call masked video autoencoder~\cite{tong2022videomae,feichtenhofer2022masked}. 
Let $\bm{v}\in\mathbb{R}^{2\tau\times C\times H\times W}$ be a short clip consisting of $2\tau$ successive frames from the entire video. %, where $\tau\in\mathbb{Z}^+$. 
A self-supervised model $f$ be formulated into an encoder-decoder framework $f(\cdot)=D(E(\cdot))$ aims to reconstruct partially masked frames in $\bm{v}$, guided by spatial and temporal relationships between tokens in adjacent frames. The encoder takes tokenized embedding vectors from a pair of successive frames using a 3D convolution~\cite{ma2022simvtp,qing2023mar,wang2023masked}. % parameterized by $\bm{w}$. 
%, $\bm{k}_i=Conv3d(\bm{v}[{2i:2i+1}];\bm{w})$.
Let $\bm{k}_i$ be an $i$-th spatiotemporal embedding vector for a pair of two frames $\bm{v}[2i:2i+1]$. Given $\bm{v}$, we reformulate the loss function as follows:
\vspace{-0.05in}
\begin{equation}\label{eq:vmim}
% \fontsize{9pt}{9pt}\selectfont
% \hspace*{-5mm}
\begin{aligned}
\ell\left(\bm{v}\right)=\sum^{\tau-1}_{i=0}
\|D\left(E\left(\bm{m}_i\otimes\bm{k}_i;\mathcal{W}_E\right);\mathcal{W}_D\right)\;\;\;\; \\
-(\bm{1}-{\bm{m}_{i}})\otimes{\bm{v}}[2i:2i+1]\|_p,\\
\end{aligned}
% &~~\text{where}~~\bm{m}=G\left(J,~\rho,~\tau\right)\in\{0,1\}^{\tau\times J},\\
% \fontsize{10pt}{10pt}\selectfont
\end{equation}
where $\bm{m}=G\left(J,~\rho,~\tau\right)\in\{0,1\}^{\tau\times J}$ and $G(\cdot)$ is a masking function depending on a specific policy, for example, random (or agnostic)~\cite{tong2022videomae} and space-only (or tube)~\cite{feichtenhofer2022masked} masking techniques.%, illustrated in \Cref{fig:concept} {(a)} and {(b)}.

% $\bm{k}_i$ indicates a $d_k$-dimensional patch embedding vector for $\bm{x}_i$, obtained by a convolutional layer parameterized by $\bm{w}$ with a stride $s$. 
$J=\left\lfloor\frac{H}{s}\right\rfloor\cdot\left\lfloor\frac{W}{s}\right\rfloor$ denotes the number of patches per image and $\|\cdot\|_p$ denotes $p$-norm. $\mathcal{W}_E$ and $\mathcal{W}_D$ are a set of weights in the encoder and decoder, respectively. $\otimes$ is a dimensionality-preserving vector-matrix multiplication operation. 
A mask $\bm{m}_{i}\in\{0,1\}^J$ is drawn by the binary distribution $B$ with a probability of $\rho$ without replacement, that is $|\bm{m}_{i}|=[J\cdot\rho]$.
After self-supervised pre-training to minimize Equation~\ref{eq:vmim}, the encoder transfers the learned representations to various downstream tasks.
Unlike the samples from the image dataset, which are permutation-invariant as they are independent of each other, consecutive frames from the video stream inherently have a strong correlation and redundancy. Thus, masked video autoencoder can enjoy spatiotemporal inductive bias from other adjacent frames in the input clip, achieving good reconstruction quality even with lessened hints (i.e., a proportion of unmasked tokens). Indeed, MVA allows a much higher masking ratio $1-\rho$ per video against MIM methods. This property is advantageous because masked modeling with a higher masking ratio significantly reduces computations when training the encoder-decoder framework.

\subsection{Challenges in Masked Video Autoencoder}\label{subsec:MVAchallenges}
MVA methods~\cite{tong2022videomae,feichtenhofer2022masked,wang2023videomae} basically capture meaningful representations from pre-training videos via random masking strategies for input tokens, which are reasonable for curated and distributionally stable video datasets, yet, there is plenty of room for further development to make the model much more robust and computation-efficient.
Here, we summarize two major \textbf{limitations} in the random sampling-based MVA: 

\textbf{(1) Patchified image tokens from a video clip are \emph{not equally} important.} At each iteration, MVA methods determine which tokens to mask according to specific random selection strategies (e.g., random, time-only, etc.). Yet, the relative amount of information in each token highly depends on the position of the informative objects and the correlation across patches within adjacent frames, which renders most of the tokens highly uninformative or redundant. These limitations lead to consuming massive training budgets in memory occupancy and suffering from slow convergence speed (e.g., training $3,200$ and $4,800$ epochs when using VideoMAE~\cite{tong2022videomae} on UCF101 and HMDB51, respectively). To address this information imbalance in visual tokens, several recent works have suggested sparsification/merging methods. EViT~\cite{liang2022not} is a supervised training method that fuses tokens by removing uninformative ones from the target task, therefore, inapplicable to video self-supervised learning. Token merging methods, including ToMe~\cite{bolya2023token}, average multiple correlated/clustered tokens and let them have the same and marginalized values using average pooling. However, the masked video autoencoder encodes a few unmasked regions of video frames and aims to reconstruct raw frames. That is, token merging techniques are inappropriate for the decoding phase of masked autoencoder by design. Furthermore, the encoding phase with token merging may also fail to reconstruct raw frames at a pixel-level, due to their marginalized token features. This technical design is well-performed in the action recognition classification problem but deteriorates the performance of the VRL method. 

\textbf{(2) MVAs draw frames at \emph{uniform time intervals} from video streams to train on.} Real-world videos may include noisy and highly redundant frame sequences, often uninformative or even detrimental in representing temporal causal relationships and features of moving objects. 
However, the uniform sampling of video frames results in a waste of computational and memory resources in video understanding. To alleviate this problem, AdaFrame~\cite{wu2019adaframe} trains the memory-augmented LSTM for gathering informative frames in given videos, leveraging supervised video labels to evaluate whether observing more frames is expected to produce more accurate predictions. SCSampler~\cite{korbar2019scsampler} also requires the ground truth to jointly train the clip classifier and clip-level saliency model to obtain useful clips to the clip classifier. SMART~\cite{gowda2021smart} trains Single-frame Selector and Global Selector for the frame selection, which require the ground truth for computing the importance score of single and paired frames.

These challenges further exacerbate the problem when applying masked video autoencoder on \textit{uncurated} first-person view real-world videos, \textit{e.g.,} Ego4D~\cite{grauman2022ego4d}, which {contains not only sparse motion information over space and time, but also suffers from a severe spatiotemporal redundancy.} Therefore, determining which frames and spatiotemporal tokens to recover is crucial for practical and efficient video representation learning. 

\section{Efficient Masked Video Autoencoder by Removing Spatiotemporal Redundancy}\label{sec:EVEREST}

\subsection{Redundancy-robust Mask Generation}\label{subsec:ours-tokenselect}
Valuable spatiotemporal information in video streams mainly comes from active visual changes rather than from static backgrounds. Thus, we aim to learn self-supervision on videos from only a few crucial regions containing minimal spatiotemporal redundancy. Let $\bm{k}_i$ be a token embedding of a pair of adjacent $2i^{th}$ and $(2i+1)^{th}$ frames from an input video clip $\bm{v}\in\mathbb{R}^{2\tau\times C\times H\times W}$, where $0\leq i< \tau$. $\bm{k}_{i+1,j}$ indicates the $j^{th}$ token embedding of $\bm{k}_{i+1}$ and we measure the importance $I_{i+1,j}$ of $\bm{k}_{i+1,j}$ by computing the \textit{distance} from the token embedding at the same region in the previous time step, $\bm{k}_{i,j}$:
\begin{equation}\label{eq:embedding}
\begin{split}    I_{i+1,j}=S\left(\bm{k}_{i+1,j},\bm{k}_{i,j}\right),\\
\end{split}
\end{equation}
where $\bm{k}_i=\texttt{Conv3d}(\bm{v}[{2i:2i+1}];\bm{w})$ and $S(\cdot,\cdot)$ indicates a distance function, such as Euclidian, negative Cosine Similarity, and negative Centered Kernel Alignment~\citep{cortes2012algorithms}. Throughout this paper, we use the $\ell_2$ norm for $S$, which is simple yet empirically performs well, and we provide a discussion for the choice of the $S$ in \highlight{Appendix B}. Tokens with a large disparity from tokens in a previous time period are considered important, indicating that they contain unique knowledge in the video and may have more important information than other tokens.
Given $\bm{v}$, the model determines the token embedding vectors with the highest importance ratio of $\rho_{pre}$ based on~\Cref{eq:embedding}, which we call \emph{Redundancy-robust (ReRo) Masking Generation}.
We can drastically reduce the computational cost during training by \textbf{only propagating these selected embedding vectors {$\widetilde{\bm{k}}$}} in a minibatch video clip at each iteration.

Next, we randomly sample a few token embeddings {$\widetilde{\bm{k}'}$} from {$\widetilde{\bm{k}}$} with the ratio of $\rho_{post}$ to forward them to the encoder. And the corresponding decoder predicts RGB pixels on all embedding vectors $\widetilde{\bm{k}}_i\in\widetilde{\bm{k}}$ based on the encoder outputs. We set a high masking ratio in general, $\rho_{pre}\cdot\rho_{post}=0.1$, to follow the settings of our MVA baselines. The objective function of our EVEREST is formulated as follows:
\begin{equation}\label{eq:patchselect}
\begin{aligned}
arg\min_{\mathcal{W},\bm{w}}~\sum^{N}_{n=1}\sum^{\tau-1}_{i=0}\left\|f_\mathcal{W}({(\widetilde{\bm{k}}_{i}')^{n}})-{{\bm{m}_{i}^{n}\otimes\bm{v}_{n}}}[2i:2i+1]\right\|_F
\end{aligned}
\end{equation}
where $\bm{m}^n=
G\left([J\cdot\rho_{pre}],\tau\right)\in\{0,1\}^{\tau\times{J}}$ and the size $|\bm{m}^n|=[J\cdot\tau\cdot\rho_{pre}]$. We omit the positional embedding term for notational simplicity. Whereas earlier MVA works used a fixed masking ratio per video cues for self-supervised training, our approach allows a dynamic masking rate for each frame by design, based on the occupancy of essential tokens. 
As shown in \Cref{fig:masking_vis}, each video frame contains a different amount of spatiotemporal information, and our dynamic masking strategy enables the model to focus on learning the valuable representation in a holistic view from incoming videos.
We note that our EVEREST can be generalized well on both pre-training and fine-tuning phases with a negligible additional computational cost and training time~(Please see \Cref{fig:resource}). We find that removing unimportant tokens is crucial for both phases, obtaining more meaningful representations and achieving superior performances. (Please see \Cref{tab:masking_ratio}). The overall process of our redundancy-robust token selection is illustrated in~\Cref{fig:main-concept-figure}. 

\begin{figure}[t]
    % \small
    \centering
    \includegraphics[width=\linewidth]{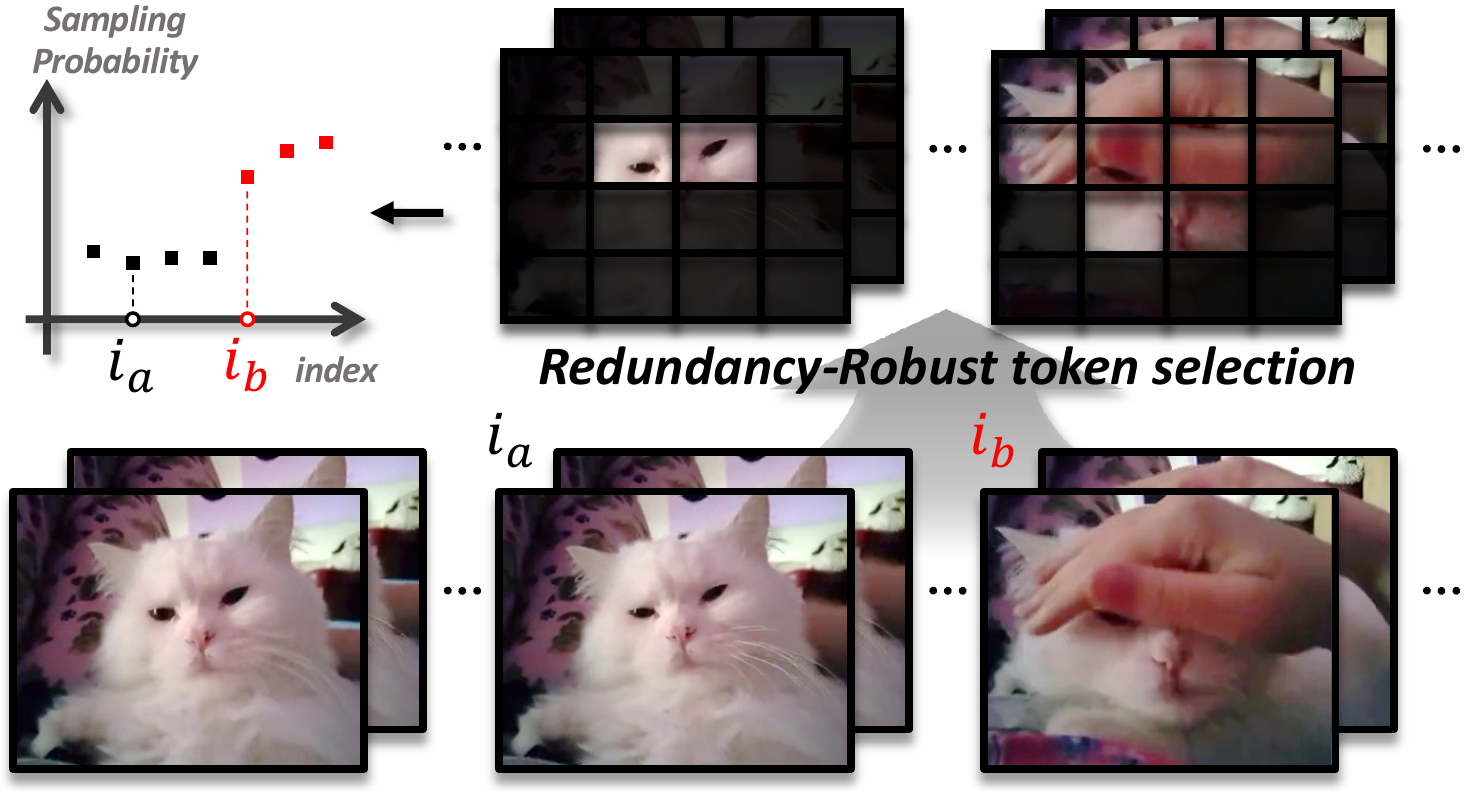}
    % \vspace{-0.1in}
     \caption{{\textbf{Information-intensive frame selection.} We adaptively select frames based on the \textbf{ReRo} token frequency, which indicates significance compared to frames.}}
    \label{fig:frame-select}
\end{figure}

\begin{figure*}
    \centering
    \includegraphics[width=0.9\linewidth]{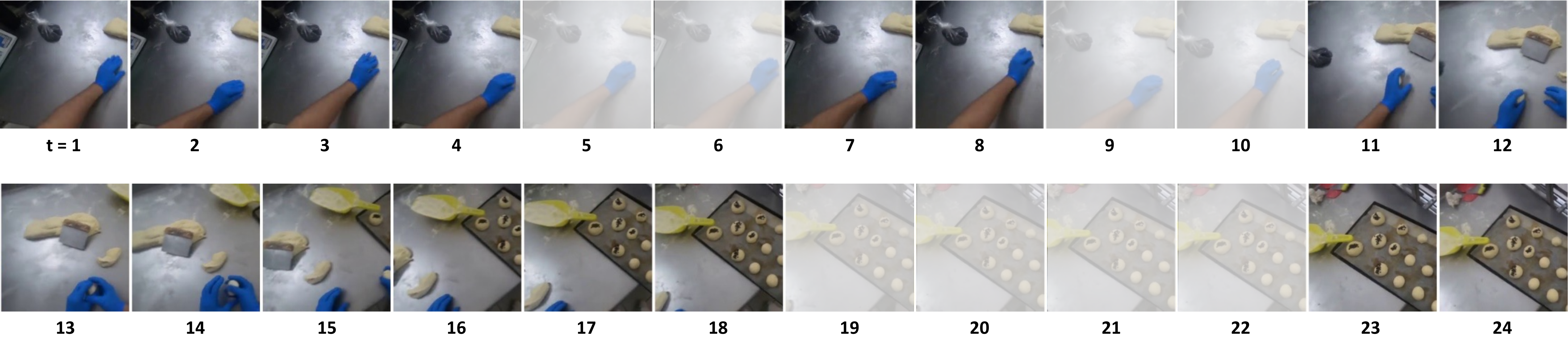}
     \caption{{\textbf{Visualization of the proposed information-intensive frame selection on an uncurated dataset, Ego4D.} Unlike prior works that uniformly samples frames similar to each other, we adaptively sample the given video (24 frames) by probabilistic sampling the frames that have distinct spatiotemporal features (non-blurred frames).}}
    \label{fig:mcs_vis}
% \vspace{-0.22in}
\end{figure*}

We emphasize that our ReRo mask generation is also significantly effective for motion-centric videos, e.g., Ego4D, since these \textit{untrimmed real-world videos also contain a lot of temporally redundant visual information}. For example, in \Cref{fig:masking-comparison}, a worker focuses on a grass mower to operate it well, and a person makes cookie dough, where visual scenes include much meaningless visual information, like empty space on the table.

\subsection{On-the-fly Information-intensive Frame Selection}\label{subsec:ours-frameselect}
As discussed in~\Cref{subsec:MVAchallenges}, real-world video streams may contain many redundant frames and also often include non-useful intermediate clips, such as temporally glancing at uninspiring walls, grounds, or skies, that interfere with estimating a causality of the user's or cameraman's attention. However, most video-based learning methods follow a simple strategy to sample frames from uniform time intervals at each iteration. 
This approach is reasonable for well-curated video benchmark datasets~\citep{kay2017kinetics,soomro2012ucf101} containing only information-dense frames with fixed viewpoints, but often unsuitable for real-world videos, which are uncurated and more likely to contain redundant and overlapped frames.

\begin{table*}[t]
\centering
% \vspace{-0.05in}
\caption{\textbf{Performance Comparison on K400.} PT and FT mean pre-training and fine-tuning, respectively. We use an input size of $16\times224^2$. Memory usage is measured with 256 batch size when pre-training. {$^\dagger$} is trained from random initialization. VideoMAE and EVEREST are pre-trained for 200 epochs, and we refer~\cite{rw2019timm} for computing {$^\ddagger$}.}
% \vspace{-0.1in}
\resizebox{\textwidth}{!}{
\begin{tabular}{l c c l l l c}
\toprule
Method      & Backbone & PT-Data & PT-GFLOPs & FT-GFLOPs & Memory usage (GB)&Top-1 Acc \\
\midrule
MViT~\citep{fan2021multiscale}{$^\dagger$}    & MViT-S    & \ding{55} &   -  &  32.9   &-&  76.0   \\
MViT~\citep{fan2021multiscale}{$^\dagger$}    & MViT-B    & \ding{55} &   -  &  70.5   &-&  78.4   \\
ViViT FE~\citep{arnab2021vivit}    & ViT-L    & IN-21K &   {119.0}\href{https://paperswithcode.com/lib/timm/vision-transformer}{$^\ddagger$}  &  3980.0   &N/A&  81.7   \\
\midrule

K-centered~\citep{Park_2022_ECCV}&{XViT}&{IN-1K}&{67.4}\href{https://paperswithcode.com/lib/timm/vision-transformer}{$^\ddagger$} &{425.0}&{N/A}&{73.1} \\
K-centered~\citep{Park_2022_ECCV}&{Mformer}&{IN-1K} &{67.4}\href{https://paperswithcode.com/lib/timm/vision-transformer}{$^\ddagger$} &{369.5}&{N/A}&{74.9} \\
K-centered~\citep{Park_2022_ECCV}&    
{TSformer}&{IN-1K} &{67.4}\href{https://paperswithcode.com/lib/timm/vision-transformer}{$^\ddagger$} &{590.0}&{N/A}&{78.0} \\

\midrule
VideoMAE~\citep{tong2022videomae}    & ViT-S    & K400 &   11.6  &  57.0   &117.4&  73.5   \\
VideoMAE~\citep{tong2022videomae}    & ViT-B    &  K400 &   35.5  &  180.5  &486.4& 78.4        \\
VideoMAE~\citep{tong2022videomae}    & ViT-L    &  K400 &  83.1  &  597.2  &634.1&  82.0   \\
\midrule
\cellcolor{gg}\textbf{EVEREST (Ours)}                         & \cellcolor{gg}ViT-S    & \cellcolor{gg}K400 &   \cellcolor{gg} \bf 6.3 \textcolor{blue}{($\downarrow45.7\%$)}   &  \cellcolor{gg}\bf 29.1 \textcolor{blue}{($\downarrow48.9\%$)}   & \cellcolor{gg}\bf 59.9 \textcolor{blue}{($\downarrow49.0\%$)} & \cellcolor{gg}\bf 75.9   \\
\cellcolor{gg}\textbf{EVEREST (Ours)}                         & \cellcolor{gg}ViT-B    & \cellcolor{gg}K400 &  \cellcolor{gg} \bf 21.5 \textcolor{blue}{($\downarrow39.4\%$)}  &  \cellcolor{gg}\bf 98.1 \textcolor{blue}{($\downarrow45.7\%$)}   &\cellcolor{gg}\bf 91.2 \textcolor{blue}{($\downarrow81.3\%$)}& \cellcolor{gg}\bf 79.2        \\
\cellcolor{gg}\textbf{EVEREST (Ours)}                         & \cellcolor{gg}ViT-L    & \cellcolor{gg}K400 & \cellcolor{gg}  \bf 60.8 \textcolor{blue}{($\downarrow26.8\%$)}  &  \cellcolor{gg}\bf 330.0 \textcolor{blue}{($\downarrow44.7\%$)}  &\cellcolor{gg}\bf 164.1\textcolor{blue}{($\downarrow74.1\%$)}&  \cellcolor{gg}\bf 82.1   \\\bottomrule
\end{tabular}}
\label{tab:vs_videomae}
% \vspace{-0.1in}
\end{table*}
To overcome the limitation, we propose to adaptively discard uninformative frames in the arrival video and build causal clips that represent the most crucial behaviors of the objects. We illustrate a simple overview of our \textit{information-intensive frame selection} in \Cref{fig:frame-select} and visualize the sampled results in \Cref{fig:mcs_vis}. 
We first select evenly spaced $[2\alpha\cdot\tau]$ frames as candidates, $\alpha$ times larger than the input clip size $\tau$, where $\alpha>1$. 
We count the number of chosen tokens $c_i$ for the $2i^{th}$ and $({2i+1})^{th}$ frames based on the frequency of our \textit{ReRo tokens}, described in~\Cref{subsec:ours-tokenselect}. Then, the model iteratively trains on the input clip by drawing $\tau$ frames from $[\alpha\cdot\tau]$ candidates without replacement, with a probability of $\frac{c_i}{\sum_i c_i}$ that the paired $2i^{th}$ and $({2i+1})^{th}$ frames are drawn. The model trains video clips with a limited length at each iteration since longer clips require massive computations and memory. Therefore, we remark that information-intensive frame selection is crucial to better capture causality in the arrival video, as the model can observe longer video fragments while avoiding redundant frames.

\begin{table*}[t!]
\begin{minipage}{.48\linewidth}
\centering
\caption{\textbf{Comparison with strong baselines on UCF101, HMDB51, and SSv2 datasets} without using the pre-training step on a large-scale dataset. Several results are drawn from \citet{diba2021vi2clr,tong2022videomae}. \textit{SR50} indicates \textit{SlowOnly-R50}.}\label{tab:ucf&hmdb}
% \vspace{-0.1in}
\resizebox{\columnwidth}{!}{
\begin{tabular}{l c c c c c}
\toprule
    \multirow{2}{*}{Method} & \multirow{2}{*}{Backbone} & \multirow{2}{*}{Extra data} & \multicolumn{3}{c}{T1 Acc~(\%)}\\
    &&&{\textbf{UCF101}}&{\textbf{HMDB51}}&{\textbf{SSv2}}\\
\midrule
    % OPN \citep{lee2017unsupervised}& {VGG} & {UCF101} & {N/A} & {59.6} & {23.8}\\
    VCOP~\citep{xu2019self}& {R(2+1)D} & {UCF101} & {72.4} & {30.9} & {N/A}\\
    CoCLR~\citep{han2020self}& {S3D-G} & {UCF101} & {81.4} & {52.1} & {N/A}\\
    Vi$^{2}$~CLR\citep{diba2021vi2clr}& {S3D} & {UCF101} & {82.8} & {52.9} & {N/A}\\
    CoCLR~\citep{han2020self}& {S3D-G} & {K400} & {87.9} & {54.6} & {N/A}\\
    Vi$^{2}$~CLR\citep{diba2021vi2clr}& {S3D} & {K400} & {89.1} & {55.7} & {N/A}\\
    RSPNet~\citep{chen2021rspnet}& {S3D-G} & {K400} & {93.7} & {64.7} & {55.0}\\    
    $\rho\text{SwAV}_{\rho=2}$~{\citep{feichtenhofer2021large}}& {SR50} & {K400}  & {87.3} &{N/A} &${51.7}$\\
    $\rho\text{MoCo}_{\rho=2}$~{\citep{feichtenhofer2021large}}& {SR50} & {K400}  & {91.0} &{N/A} &${54.4}$\\
    $\rho\text{BYOL}_{\rho=2}$~{\citep{feichtenhofer2021large}}& {SR50} & {K400}  & {92.7} &{N/A} &${55.8}$\\
% \midrule    
    VideoMAE~\citep{tong2022videomae}&{ViT-B}&{\ding{55}} &{91.3} & {62.6} & {64.3}\\
\midrule
    \cellcolor{gg}\textbf{EVEREST (Ours)}&\cellcolor{gg}{ViT-B}&\cellcolor{gg}{\ding{55}}& \cellcolor{gg}\textbf{93.4} & \cellcolor{gg}\textbf{68.1} & \cellcolor{gg}\textbf{64.6}\\
\bottomrule
\end{tabular}}
\end{minipage}\hspace{0.15in}
\begin{minipage}{.48\linewidth}
\centering
    \includegraphics[width=\linewidth]{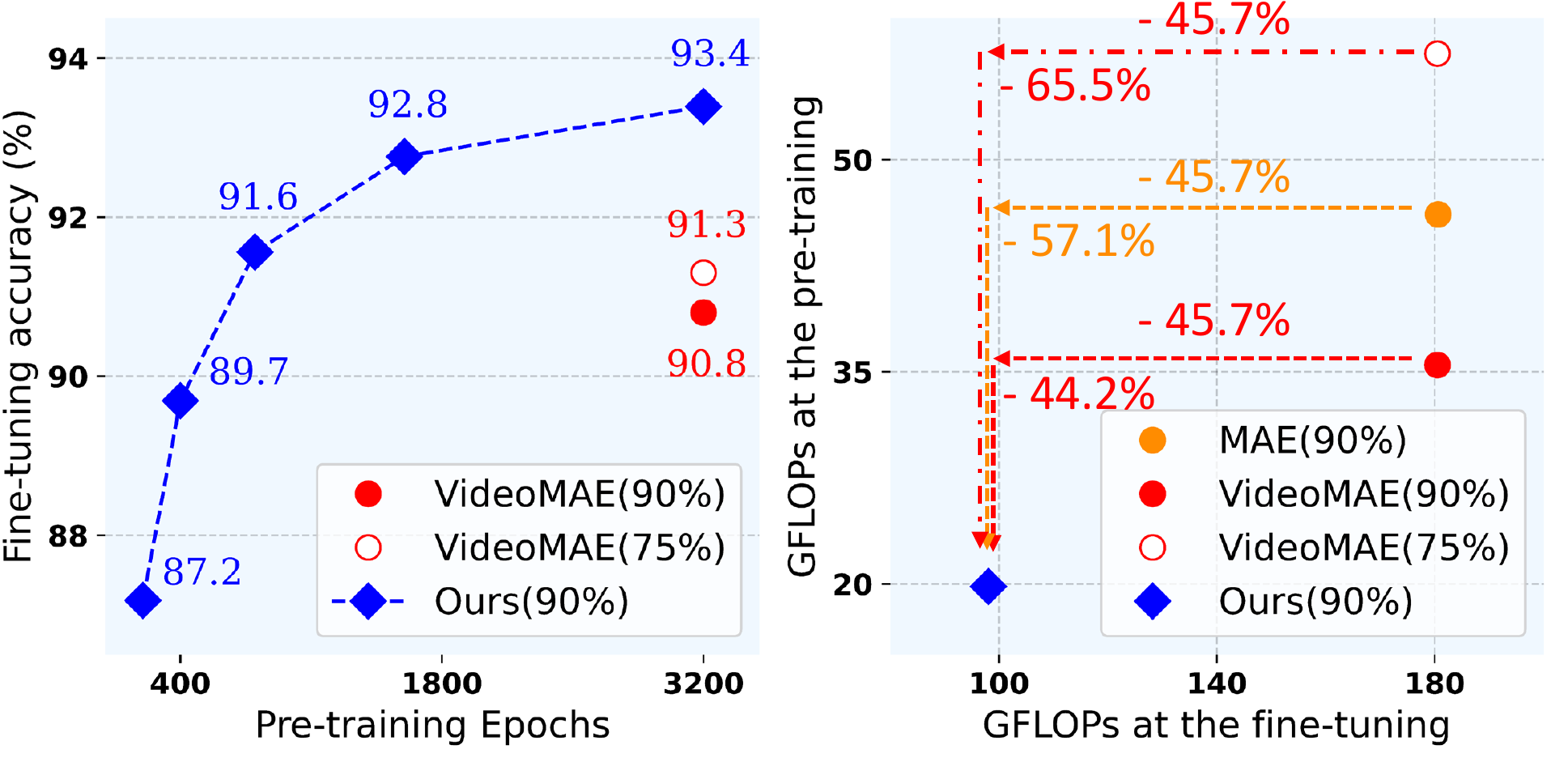}
    % \vspace{-0.1in}
    \captionof{figure}{ \textbf{Performance \& GFLOPs Comparison} on UCF101 dataset. \textbf{(Left)} EVEREST outperforms VideoMAE for both masking ratios (75\% and 90\%) even at significantly fewer training epochs. %We additionally report the results of VideoMAE with a lower masking ratio (75\%) during pre-training. 
    \textbf{(Right)} Our EVEREST reduces GFLOPs during pre-training and fine-tuning compared to VideoMAE and ST-MAE.}
    \label{fig:a}

\end{minipage}
\end{table*} 

\section{Experiments}\label{experiments}
\paragraph{Experimental settings}
\label{Experimental Settings}
We validate our method on multiple video datasets:  
\emph{UCF101}~\cite{soomro2012ucf101}, \emph{HMDB51}~\cite{kuehne2011hmdb}, 
\emph{Something-Something v2 (SSv2)}~\cite{goyal2017something}, 
\emph{Kinetics-400~(K400)}~\cite{kay2017kinetics} and \emph{Ego4D}~\cite{grauman2022ego4d}.  
% {UCF101}, {HMDB51}, {SSv2} and {K400} are mainly used datasets for video action recognition. 
We evaluate our information-intensive frame selection strategy during pre-training on \emph{Object State Change Classification (OSCC)}%\footnote{https://github.com/EGO4D/hands-and-objects/tree/main/state-change-localization-classification} 
 task from {Ego4D}, containing raw and uncurated people’s daily life videos. Given 8-second videos, OSCC classifies whether the object’s state has changed by interacting with a camera wearer. 
Following the same training protocols as VideoMAE~\cite{tong2022videomae}, we pre-train our EVEREST over benchmark datasets \emph{without labels} and report the fine-tuning performance.
For a fair comparison, we train both VideoMAE and our EVEREST using the same \emph{one-node} equipped with 8 GPUs. For K400 and SSv2, although VideoMAE trained all models for 1600 and 2400 epochs, respectively, with multi-node GPUs~(e.g., 64 V100 GPUs), we train both methods with several scaled ViT backbones for 200 epochs due to quick validation of the scalability. Except for K400, we use ViT-B as a backbone for the other benchmarks. Please see \highlight{Appendix A} for further implementation details.

\textbf{EVEREST is \evblue{memory and computationally efficient} while achieving competitive performance} against VideoMAE and K-centered variants. 
We extensively compare our proposed method against strong VRL baselines. 
\Cref{tab:vs_videomae} shows the results of the supervised and self-supervised methods on K400.
Our redundancy-robust token selection, EVEREST, shows comparable performance to VideoMAE~\cite{tong2022videomae}, while achieving \textit{significant} computation and memory reduction during both pre-training and fine-tuning.
Specifically, for ViT-S, EVEREST can reduce computational costs by 45.7\% and 48.9\%, respectively.
It's also worth noting that EVEREST using ViT-L is more than $4\times$ more memory efficient than VideoMAE~(e.g., 164.1 GB vs. 634.1 GB).
Meanwhile,~\Cref{tab:ucf&hmdb} shows that EVEREST achieves superior fine-tuning performance against recent VRL methods across UCF101 and HMDB.
Specifically, EVEREST outperforms the best baseline, VideoMAE, by $2.1\%p\uparrow$ on UCF101 and $3.2\%p\uparrow$ on HMDB51. Also, we visualize the convergence plot of EVEREST on UCF101 in~\Cref{fig:a} \highlight{Left}. By pre-training at only 800 epochs, our EVEREST surpasses the fine-tuning accuracy of VideoMAE trained at 3200 epochs. That is, EVEREST reaches on-par performance with VideoMAE by using only \textbf{$\sim$14\% of total computational costs}. 
We also against variants of a strong motion-based token selection method, K-centered patch sampling, with the modified vision transformer for video learning, including XViT\cite{bulat2021space}, Mformer~\cite{patrick2021mformer}, and TSformer~\cite{bertasius2021timesformer}. Our proposed method also surpasses these motion-based video understanding methods by significant margins in terms of Top-1 accuracy and GFLOPs.

\begin{table*}[t!]
\centering
\begin{minipage}{.48\linewidth}
    \caption{\textbf{Memory usage comparison} during the pre-training (one node, A100$\times$8 GPUs) on K400. The memory gaps grow up when increasing batch size and model size. %Especially, our EVEREST with ViT-L achieves about $\mathbf{4\times}$ \textbf{better} efficiency than VideoMAE (634.1GB). 
    \highlight{\emph{OOM}} indicates \textit{out-of-memory}.}
    \vspace{-0.1in}
    \center
    % \begin{adjustbox}{width=\columnwidth, center}
    \resizebox{0.9\linewidth}{!}{
    \begin{tabular}{cccccc}
    \toprule
        \multirow{2}{*}{Method} & \multirow{2}{*}{Backbone} & \multicolumn{4}{c}{Effective Batch Size}\\
        && 128 & 256 & 512 & 1024\\
    \midrule
    \vspace{0.03in}
    &{ViT-S} & 71.9  & 117.4  & 233.6  & 443.6  \\ 
    \vspace{0.03in}
    VideoMAE&{ViT-B} & 258.4  & 486.4  & \highlight{\emph{OOM}} & \highlight{\emph{OOM}}\\
    \vspace{0.03in}
    &{ViT-L} & 353.9  & 634.1  & \highlight{\emph{OOM}} & \highlight{\emph{OOM}}\\
    \midrule
     % \vspace{0.03in}
    \multirow{3}{*}{\begin{tabular}{c}\textbf{EVEREST}\\\textbf{(Ours)}\end{tabular}}  &{ViT-S}& \textbf{31.9 } & \textbf{59.9 } & \textbf{95.6 } & \textbf{173.1 }\\
    \vspace{0.03in}
    &{ViT-B}& \textbf{60.0 } & \textbf{91.2 } & \textbf{151.2 } & \textbf{273.6 }\\
    \vspace{0.03in}
    &{ViT-L}& \textbf{117.3 } & \textbf{164.1 } & \textbf{258.9 } & \textbf{428.4 }\\
    \bottomrule
    \end{tabular}}
    % \vspace{-0.2in}
    \label{tab:mem-allocation2}

\end{minipage}\hspace{0.15in}
\begin{minipage}{.48\linewidth}
  \caption{\textbf{Pre-training time \& Memory comparison with SoTA MVAs on K400.} We measure the pre-training time for an epoch under a single-node machine equipped with 8$\times$A6000~(48GB) GPUs. We use ViT-B and a batch size of 128. Note that MME~\citep{sun2023mme} should pre-compute the optical flow for the entire video data before pre-training. We exclude the time and memory of MME for optical flow computations in the table.}
    \center
    \resizebox{0.98\linewidth}{!}{
        \begin{tabular}{lcc}
        \hline
        Method                 & PT-Time & Memory \\ \hline
        VideoMAE~\citep{tong2022videomae}               & 18m   42s            & 150.3 GB              \\
        MME~\citep{sun2023mme}                    & 10m   15s            & 121.2 GB              \\
        MVD~\citep{wang2023mvd}                    & 51m   55s            & 274.9 GB              \\
        \cellcolor{gg}\textbf{EVEREST   (Ours)} & \cellcolor{gg}\textbf{8m   18s}    & \cellcolor{gg}~\textbf{66.3} GB      \\ \hline
        \end{tabular}
    }    
  \label{tab:pt_mem}%
\end{minipage}\\
\vspace{0.15in}
\begin{minipage}{.48\linewidth}

    \caption{\textbf{EVEREST-Finetuning with other MVAs on K400.} We apply our EVEREST for finetuning the pre-trained models~(ViT-B) by SoTA MVAs. We measure the memory usage with the same batch size of 128. While SoTA methods use full tokens during fine-tuning, EVEREST uses only redundancy-robust tokens.}
    % \vspace{-0.15in}
    \center
    \resizebox{\linewidth}{!}{
        \begin{tabular}{ccccc}
        \hline
        PT-Method & FT-Method       & GFLOPs & Memory  & Top-1         \\ \hline
        VideoMAE  & Full-token      & 180.5  & 362.5 GB      & 81.5          \\
        \cellcolor{gg}VideoMAE  & \cellcolor{gg}\textbf{EVEREST}   & \cellcolor{gg}98.1   & 1\cellcolor{gg}78.4 GB      & \cellcolor{gg}81.6          \\ \hline
        MME       & Full-token      & 180.5  & 362.5 GB      & 81.8          \\
        \cellcolor{gg}MME       & \cellcolor{gg}\textbf{EVEREST}   & \cellcolor{gg}98.1   & \cellcolor{gg}178.4 GB      & \cellcolor{gg}82.0          \\ \hline
        MVD       & Full-token      & 180.5  & 362.5 GB      & 83.4          \\
        \cellcolor{gg}MVD       & \cellcolor{gg}\textbf{EVEREST}   & \cellcolor{gg}98.1   & \cellcolor{gg}178.4 GB      & \cellcolor{gg}82.8          \\ \hline
        \end{tabular}
    }
    \label{tab:ft_mem}
  % \vspace{-0.4cm}
\end{minipage}\hspace{0.15in}
\begin{minipage}{.48\linewidth}
  
\centering
    % \vspace{-0.05in}
    \includegraphics[width=\linewidth]{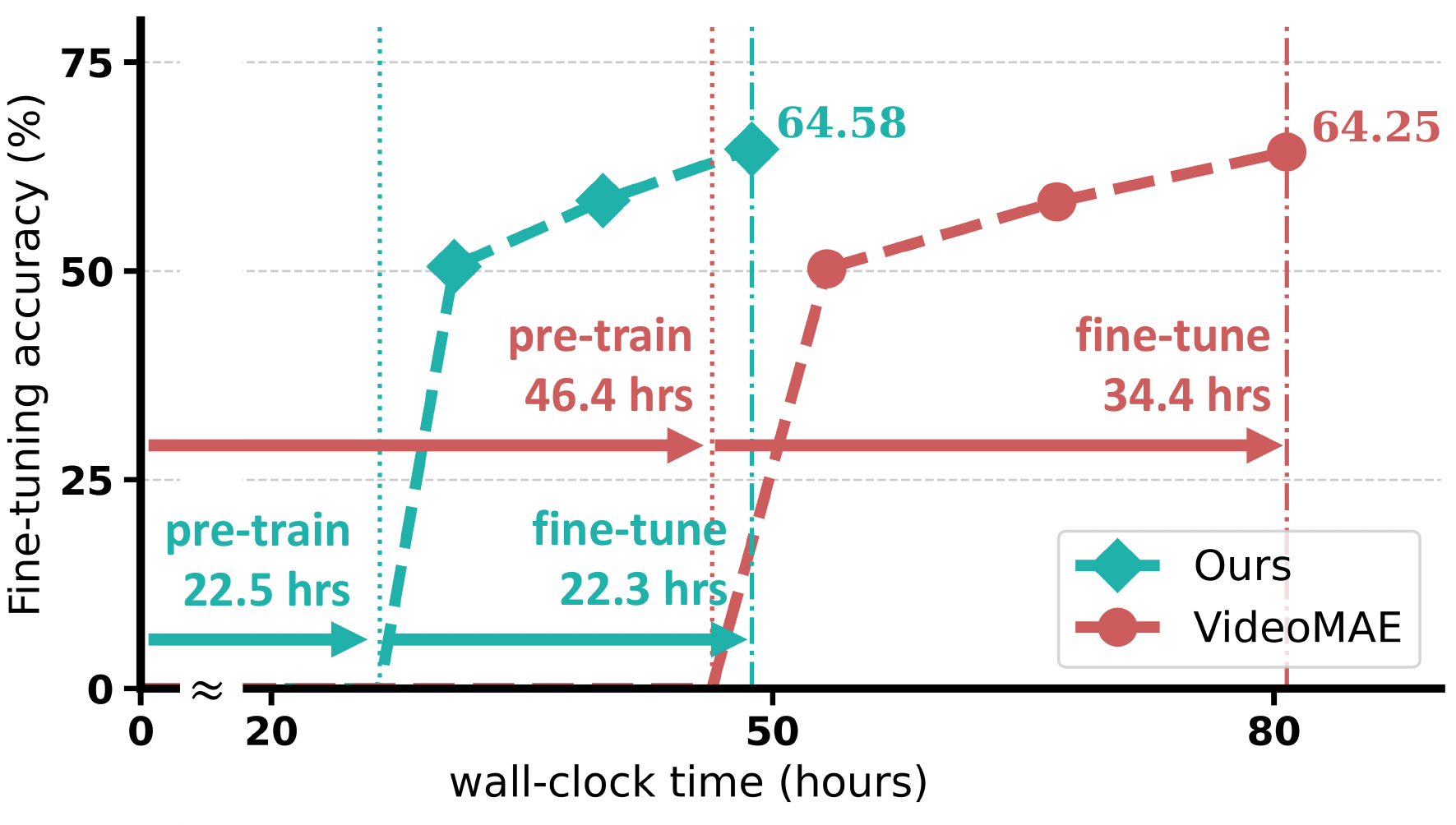}
    % \vspace{-0.1in}
     \captionof{figure}{\textbf{Accuracy over training time} on SSv2. 4 NVIDIA RTX 3090 GPUs are used. We set $\rho_{pre}$ to 0.3 and 0.8 for pre-train and fine-tuning, respectively.}
    \label{fig:ssv2_time}
    % \vspace{-0.15in}
\end{minipage}
\end{table*}

\textbf{EVEREST is highly beneficial for \evblue{model deployment}.}
To validate the efficiency of our EVEREST in terms of memory usage, we measure the memory consumption over multiple batch sizes and architecture scales during the pre-training phase on K400.
We use one node equipped with $8\times$A100 (80GB) GPUs and compare EVEREST with VideoMAE.
As shown in~\Cref{tab:mem-allocation2}, VideoMAE shows \textbf{\emph{out-of-memory}} when using a batch size of 512 and 1,024. Regarding that VideoMAE used the batch size of 1,024 in the original paper~\cite{tong2022videomae}, VideoMAE inevitably has to lower the batch size to run training in a one-node environment, resulting in increased training time.
However, the proposed redundancy-robust token selection approach allows to drastically reduce the memory usage, which means that it can be trained using only a one-node environment. We further compare pre-training and fine-tuning budgets with those of the state-of-the-art MVAs such as MME~\cite{sun2023mme} and MVD~\cite{wang2023mvd}.
In~\Cref{tab:pt_mem}, when pre-training, EVEREST requires only 44\%, 24\%, and 55\% of memory than VideoMAE, MME and MVD, respectively.
Due to the generality of our EVEREST, we can apply EVEREST to the finetuning phase by prunning redundant tokens. As shown in~\Cref{tab:ft_mem}, compared to other methods using full tokens, our EVEREST reducing computation and memory requirements while acheiving comparable accuracy.

\textbf{EVEREST is also significantly \evblue{rapid} for pre-training compared to SoTA VMA baselines.}
To evaluate the wall-clock time efficiency, we compare pre-training time~(PT-Time) with state-of-the-art MVAs by using the same batch size on K400. As shown in~\Cref{tab:pt_mem}, EVEREST requires only 45\%, 16\%, and 81\% of pre-training time than VideoMAE, MME, and MVD, respectively. Note that we didn't reflect the heavy computation burdens of MME for extracting HOG and optical flow of all input video data before training. Similarly, the whole training time, including both pre-training and fine-tuning, can be drastically reduced, as shown in~\Cref{fig:ssv2_time}. 
% For a fair comparison, we use the largest batch sizes that can fit into the memory during pre-train and fine-tuning, which are 256/48 and 96/32 for ours and VideoMAE, respectively.
In short, besides a gain in memory efficiency, we can maximally enjoy remarkably accelerated pre-training and fine-tuning in resource-constrained environments compared to existing MVA methods.

\begin{figure*}
    \small
    \center
    \resizebox{\linewidth}{!}{%
    \begin{tabular}{cccc}
    \includegraphics[width=0.2\linewidth]{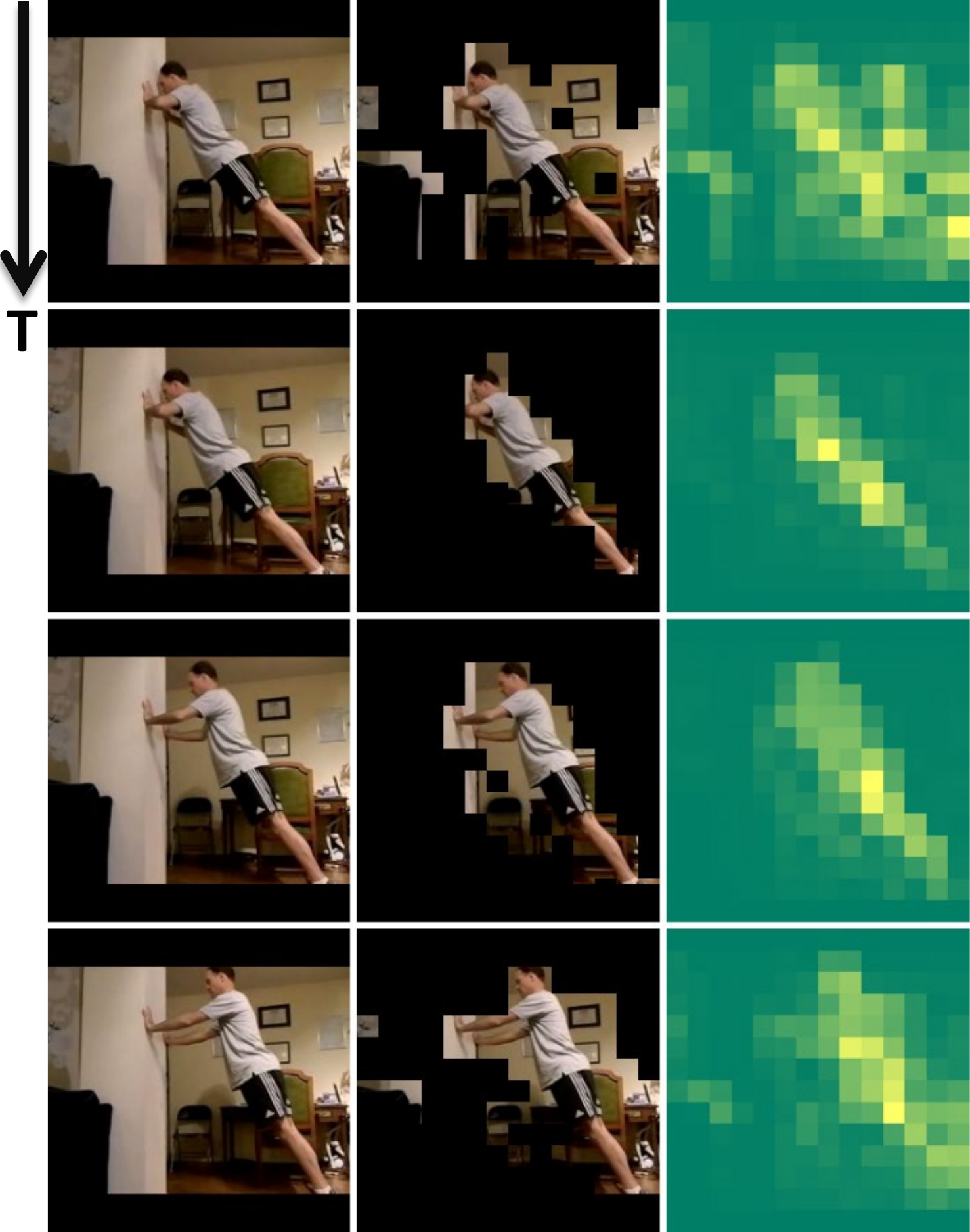}&
    \includegraphics[width=0.2\linewidth]{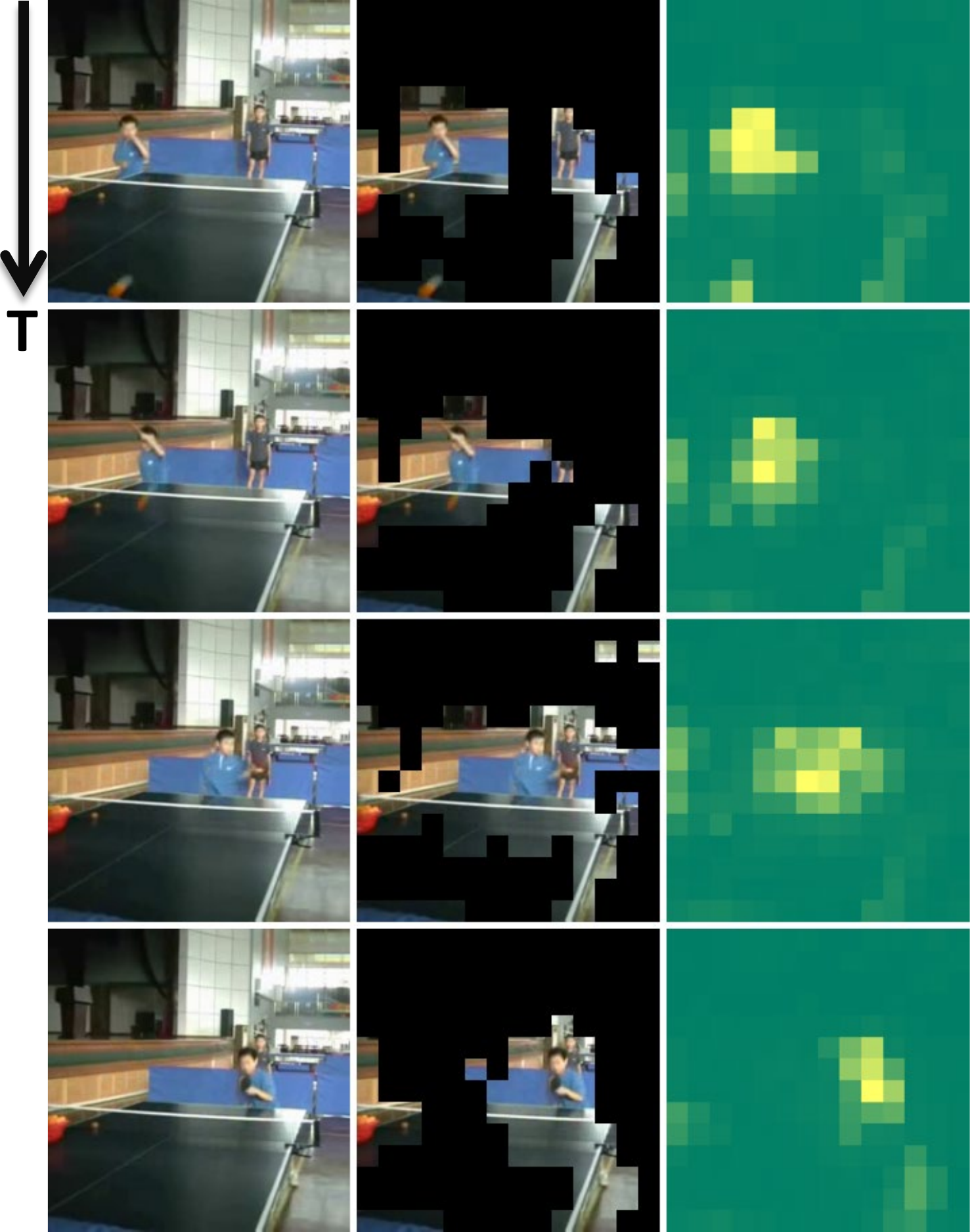}&
    \includegraphics[width=0.2\linewidth]{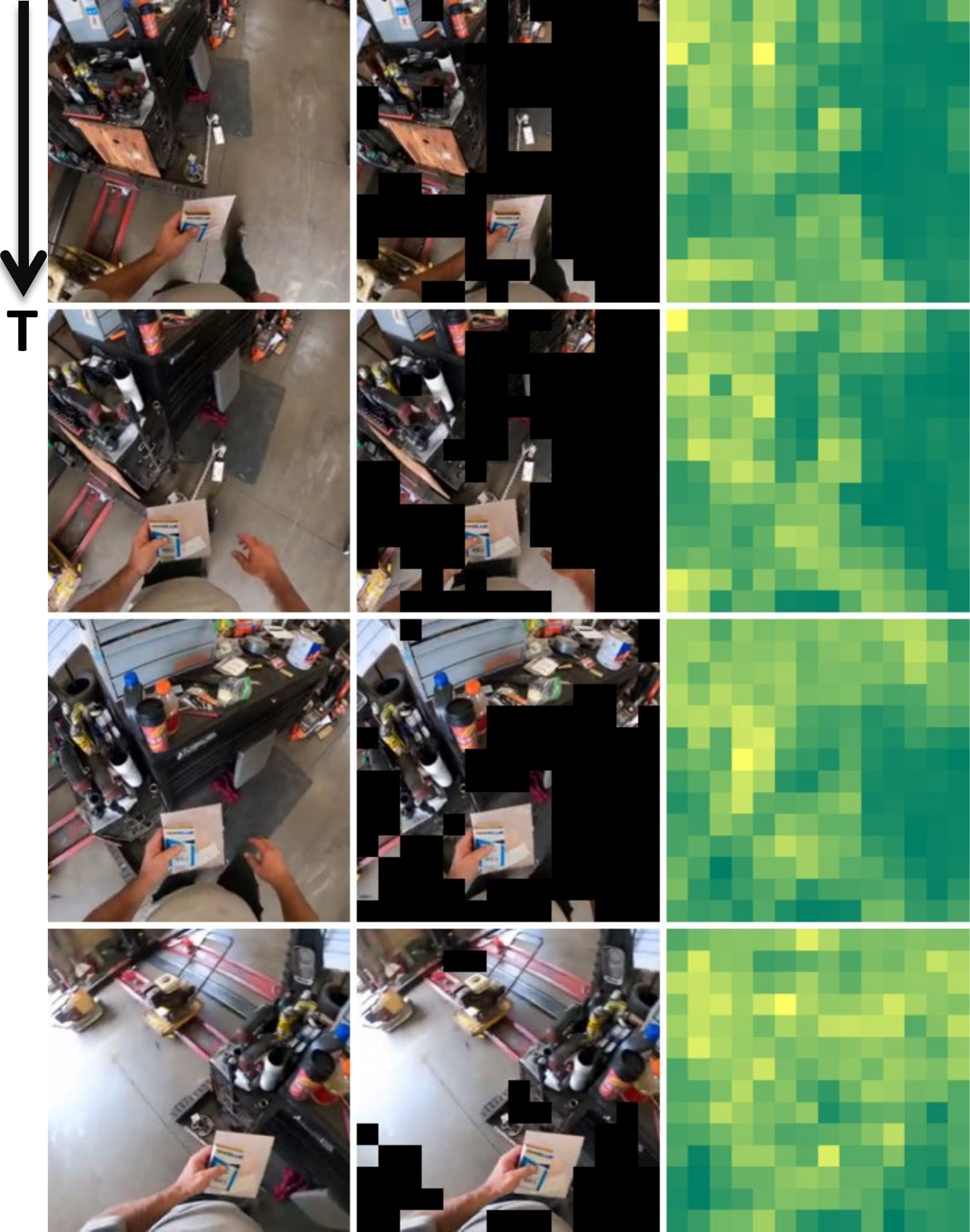}&
    \includegraphics[width=0.2\linewidth]{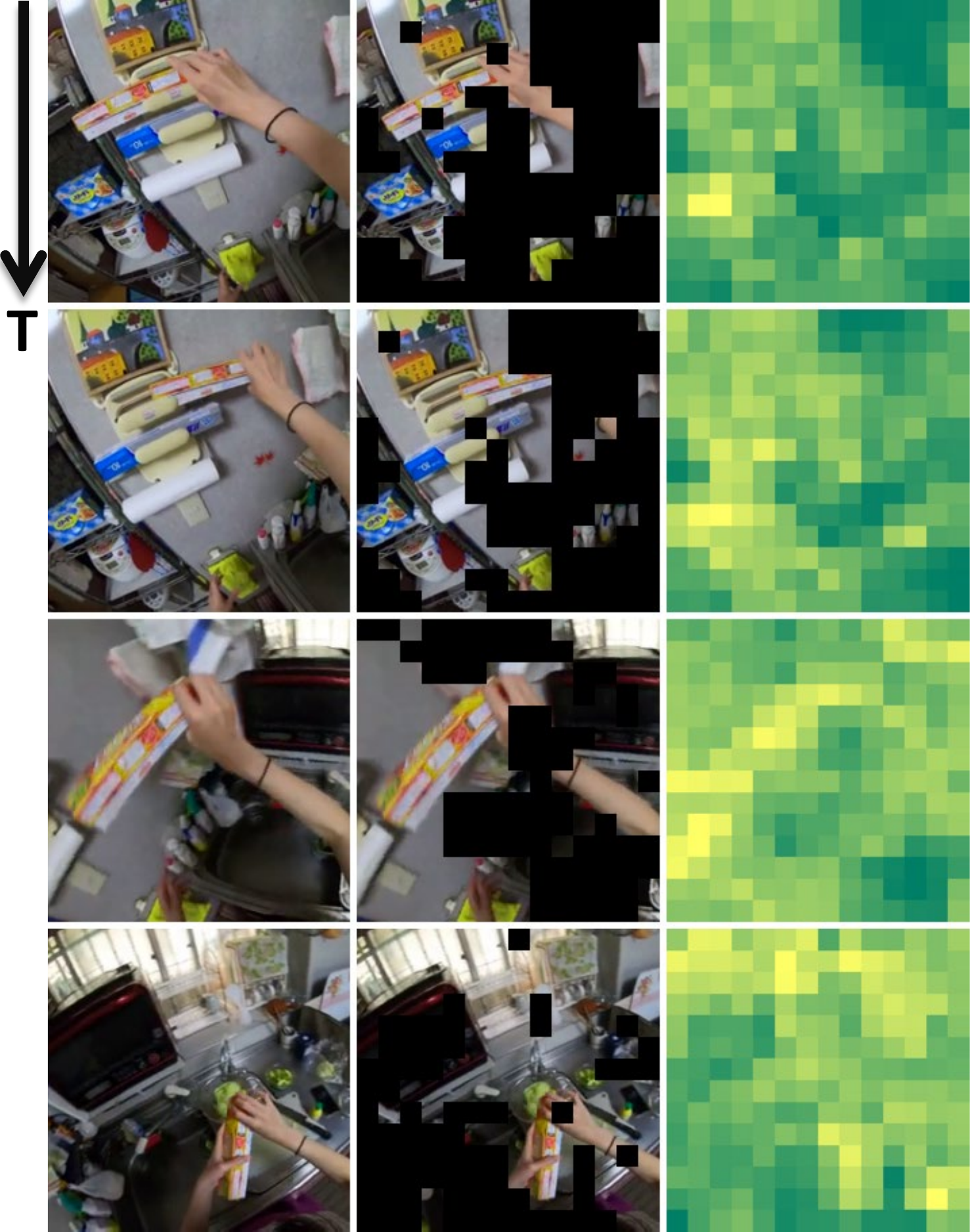}\\
         \textbf{(a) fixed and static}&
         \textbf{(b) fixed and dynamic}&
         \multicolumn{2}{c}{\textbf{(c) first-view egocentric videos}}
% \vspace{0.05in}\\
    \end{tabular}}
% \vspace{-0.1in}
 \caption{{\textbf{Examples of Redundancy-Robust (ReRo) Token Selection.} We visualize the raw video frames (left), Redundancy-Robust masking (middle), and obtained importance heatmaps (right) on UCF101 (\textbf{(a)} and \textbf{(b)}, $\rho_{pre}$=$0.3$) and Ego4D (\textbf{(c)}, $\rho_{pre}$=$0.5$). Our ReRo token selection successfully captures narrow/concentrated motion information \textbf{(a, b)} and distributed/multi-object motion cues \textbf{(c)}.}}
    \label{fig:masking_vis}
% \vspace{-0.15in}
\end{figure*}

\textbf{EVEREST has a potential for estimating \evblue{motion information}.}
Our method is also effective for egocentric videos, e.g., Ego4D, which capture dynamic views from camera motion. EVEREST captures informative objects even with the rapid change of view in first-person videos, as it masks out objects crucial to understanding the camera wearer's attention while not attending to the background. Specifically, we provide masked input examples with their importance heatmaps through our proposed redundancy-robust masking in~\Cref{fig:masking_vis}. Our masking strategy enables the model to capture the most informative tokens containing static and dynamic moving of objects (e.g., \emph{doing wall push up} (\Cref{fig:masking_vis} \highlight{(a)}) and \emph{playing ping pong} (\Cref{fig:masking_vis} \highlight{(b)}). In addition, it captures informative objects even containing drastic viewpoint changes in videos, as it masks objects crucial for understanding the camera wearers’ attention~(\Cref{fig:masking_vis} \highlight{(c)}), while not focusing on backgrounds such as walls and floors.

Furthermore, it is worth noting that egocentric videos contain temporally redundant visual information. \Cref{fig:masking-comparison} reveals that a worker focuses on a grass mower to operate it well, and a person makes cookie dough, where visual scenes include much meaningless visual information, such as empty space on the table. Our method automatically discards these noisy and redundant tokens in an online manner without requiring additional dense information, such as optical flows of incoming videos.

These abundant redundant scenes over a long time in real-world videos market VRL models to waste time and computations by learning on meaningless frames, which also can lead to poor local optima due to bias and catastrophic forgetting. To overcome the limitation, we adopt the redundancy robust sampling rate $\alpha$ to focus on frames with larger motions across frames (See~\Cref{fig:mcs_vis}).
By constructing the given videos with fluent motion information, our model focuses more on learning the core part of the video. We set the default $\alpha$ to 1.5 and report the effect of $\alpha$ in~\highlight{Appendix C}. %\Cref{sec:supple:Additional Experimental Results}}. 
We quantitatively compare our method with recent SoTA methods using the OSCC task on Ego4D in~\Cref{tab:table_oscc_results}. We pre-train OSCC without labels and fine-tune it to classify whether the object's state has changed. With only visual information, we outperform the previous SoTA method, Egocentric VLP~\cite{kevin2022egovlp}, which uses visual and text information, by $2.3\%p\uparrow$.

\begin{table}[t]
    % \vspace{-0.05in}
    \caption{\textbf{Comparison with public SoTA methods for OSCC on Ego4D val set.} We set our frame selection ratio $\alpha=1.5$. Pre-trained data modalities from the OSCC dataset ‘V’ and ‘T’ refers to visual and text, respectively.}
    % \vspace{-0.1in}
    \center
    \resizebox{\linewidth}{!}{
        \begin{tabular}{lcl}
        \toprule
        Method         & Modality & Accuracy~(\%)      \\ \midrule
        Egocentric VLP~\citep{kevin2022egovlp} & V+T      & 73.9          \\ \midrule
        SViT~\citep{escobar2022video}           & V        & 69.8          \\
        TarHeels~\citep{islam2022object}       & V        & 70.8          \\
        \cellcolor{gg}\textbf{EVEREST (Ours)} & \cellcolor{gg}V & \cellcolor{gg}{\textbf{76.2} \textbf{\footnotesize\highlight{(+5.4\%p)}}} \\ \bottomrule
        \end{tabular}
    }
    % \vspace{-0.1in}
    \label{tab:table_oscc_results}
% \vspace{-5mm}
\end{table}
\section{Conclusion}\label{conclusion}
From the insight that not all video tokens are equally informative, we propose a simple yet efficient parameter-free token and frame selection method for video pre-training. We select redundancy-robust tokens based on their significance in the change of their states and train the model only with them, drastically reducing memory allocations and computations. In addition, we propose a frame selection technique to construct input video data by sampling frames with a probability proportional to the degree of their occupancy of selected tokens. This is useful for video representation learning with uncurated videos containing a number of redundant frames. 
We empirically demonstrate that our method is significantly more efficient in computations, memory, and training time than strong baselines.
We believe our method could help democritizing~\cite{seger2023democratising} video-related research requiring immense computation budgets.

% \section*{Broader impact}
% In this work, we propose EVEREST and compare it with strong video representation learning baselines on multiple benchmarks. Our goal is to improve training/inference efficiency in video understanding. Our work could have potential societal consequences, none of which we feel must be specifically highlighted here.

% the National Research Foundation of Korea(NRF) grant funded by the Korea government(MSIT) (NO. RS-2023-00256259). 리더연

% \section{Acknowledgement}
% This work is supported by
\section*{Impact Statement}\label{sec:supple:impacts_and_limitations}
Our work has a positive broader societal impact leading to environment-friendly AI research by drastically reducing energy and carbon costs in the training and inference phases. Current methods in the field of video understanding consume enormous monetary and environmental costs, and we believe that our work contributes to overcoming this limited feasibility of video-based research fields by enabling single-node training with larger batch sizes using a large backbone, whereas existing VRL methods require immense memory occupancy and suffer from deploying models in the same setup due to \textit{out-of-memory}. We also aim to enhance the accessibility of those cutting-edge video models by employing practical resources on a single node and pursuing eco-friendly AI research.

% However, while our EVEREST achieves impressive performance improvements along with drastic reduction in memory, computation, and training time over multiple benchmark datasets/tasks with different backbone sizes, the improvement in model accuracy for the action recognition task on the K400 dataset becomes relatively small for huge backbones such as ViT-L. We speculate that this is due to their better ability to encode spatiotemporal representations into higher-dimensional latent spaces. In addition, our method focuses on capturing short-term motion changes in videos by measuring the temporal disparity between adjacent frames. Although we validated the effectiveness of our method on egocentric videos focusing on temporal semantics or causal interactions, the effect of our method on tasks of long-term episodic memory or the cameraman's intention/interactions under a long horizon view was not thoroughly analyzed in this study. We leave it to future research to analyze and address these limitations.

\section*{Acknowledgement}
This work was supported by Institute for Information \& communications Technology Promotion (IITP) grant funded by the Korea government (MSIP) (No.2019-0-00075 Artificial Intelligence Graduate School Program (KAIST)), Institute of Information \& communications Technology Planning \& Evaluation (IITP) grant funded by the Korea government (MSIT) (No. RS-2022-00187238, Development of Large Korean Language Model Technology for Efficient Pre-training), the National Research Foundation of Korea (NRF) grant funded by the Korea government (MSIT) (NO. RS-2023-00256259), KAIST-NAVER Hypercreative AI Center, Google Research Grant, and Google Cloud Research Credits program with the award (KX3L-GHDM-4U6F-QLE0).
% GCP 추가해야됨
%------------------------------------------------------------------------

% \bibliographystyle{plain}
\bibliography{egbib}
% \bibliography{iclr2024_conference}
\bibliographystyle{icml2024}

\clearpage
\appendix

\paragraph{\LARGE{Appendix}}

\paragraph{Organization} The supplementary file is organized as follows: Firstly, we explain the implementation details according to tasks in~\Cref{sec:supple:Implementation Details} and perform the distance function ablation study for redundancy-robust masking during pre-training in~\Cref{sec:supple:dist_function}. We provide additional experimental results for our EVEREST in~\Cref{sec:supple:Additional Experimental Results}. We visualize more examples from EVEREST in~\Cref{sec:supple:Visualization of motion-centric masking and sampling}.

\section{Implementation Details \label{sec:supple:Implementation Details}}
\begin{table}[h]
\centering
% \vspace{-0.1in}
%\addtolength{\tabcolsep}{-2.8pt}  
\caption{\textbf{Pre-training settings for K400, SSv2, UCF101, HMDB51 and OSCC.}}
% \vspace{-0.15in}
\resizebox{\columnwidth}{!}{
\renewcommand{\arraystretch}{1}
\renewcommand{\tabcolsep}{1pt}
\begin{tabular}{c|ccccc}

& {K400} & {SSv2} & {UCF101} & {HDMB51} & {OSCC}\\
\midrule
optimizer&\multicolumn{5}{c}{AdamW}\\
optimizer&\multicolumn{5}{c}{\multirow{2}{*}{$\beta_1, \beta_2=0.9, 0.95$~\cite{chen2020generative}}}\\
momentum&\\
\multirow{2}{*}{base learning rate}&{1.5e-4(S,L),}&\multirow{2}{*}{1.5e-4}&\multirow{2}{*}{1e-3}&\multirow{2}{*}{1e-3}&\multirow{2}{*}{1e-3}\\
&{3e-4(B)}&&&&\\
weight decay&\multicolumn{5}{c}{0.05}\\
learning rate schedule&\multicolumn{5}{c}{Cosine decay~\cite{loshchilov2016sgdr}}\\
flip augmentation&{yes}&{no}&{yes}&{yes}&{yes}\\
augmentation&\multicolumn{5}{c}{MultiScaleCrop}\\
\midrule
\multirow{2}{*}{batch size}&{1024(S,B),}&\multirow{2}{*}{256}&\multirow{2}{*}{192}&\multirow{2}{*}{192}&\multirow{2}{*}{256}\\
&{512(L)}&&&&\\
warmup epochs&{40}&{40}&{40}&{40}&{20}\\
$\rho_{pre}$&{0.3}&{0.3}&{0.3}&{0.3}&{0.5}\\
sampling stride&{4}&{2}&{4}&{2}&{4}\\
total epochs&{200}&{200}&{3200}&{4800}&{400}\\

\end{tabular}}
\label{tab:hyperparams_ucf_hmdb_pt}
% \vspace{-0.1in}
\end{table}
\begin{table}[h]
\centering
% \vspace{-0.1in}
\caption{\textbf{Fine-tuning settings for K400, SSv2, UCF101, HMDB51 and OSCC.}}
% \vspace{-0.15in}
%\addtolength{\tabcolsep}{-2.8pt}  
\resizebox{\columnwidth}{!}{
\renewcommand{\arraystretch}{1.0}
\renewcommand{\tabcolsep}{1pt}
\begin{tabular}{c|ccccc}

&{K400} & {SSv2} & {UCF101} & {HDMB51} & {OSCC}\\
\midrule
optimizer&\multicolumn{5}{c}{AdamW}\\
optimizer&\multicolumn{5}{c}{\multirow{2}{*}{$\beta_1, \beta_2=0.9, 0.999$}}\\
momentum&\\
weight decay&\multicolumn{5}{c}{0.05}\\
learning rate schedule&\multicolumn{5}{c}{Cosine decay}\\
warmup epochs&\multicolumn{5}{c}{5}\\
laer-wise lr decay&\multicolumn{5}{c}{0.75~\cite{bao2021beit}}\\
flip augmentation&{yes}&{no}&{yes}&{yes}&{yes}\\
RandAug&\multicolumn{5}{c}{$(9, 0.5)$~\cite{cubuk2020randaugment}}\\
label smoothing&\multicolumn{5}{c}{0.1~\cite{szegedy2016rethinking}}\\
drop path&\multicolumn{5}{c}{0.1}\\
\midrule
\multirow{3}{*}{base learning rate}&{5e-4(S),}&\multirow{3}{*}{5e-4}&\multirow{3}{*}{1e-3}&\multirow{3}{*}{1e-3}&\multirow{3}{*}{1e-4}\\
&{1e-3(B),}&&&&\\
&{2e-3(L)}&&&&\\
\multirow{2}{*}{batch size}&{384(S,B),}&\multirow{2}{*}{48}&\multirow{2}{*}{128}&\multirow{2}{*}{128}&\multirow{2}{*}{32}\\
&{128(L)}&&&&\\
$\rho_{pre}$&{0.6}&{0.8}&{0.6}&{0.6}&{-}\\
sampling stride&{4}&{-}&{4}&{2}&{10}\\
total epochs&{75(S), 35(B,L)}&{50}&{100}&{50}&{30}\\
multi-view&{2$\times$3}&{2$\times$3}&{5$\times$3}&{10$\times$3}&{2$\times$3}\\

% \hline \hline

\end{tabular}}
\label{tab:hyperparams_ucf_hmdb_ft}
\end{table}
% \vspace{-0.1in}

We validate our EVEREST on five video datasets: \emph{K400}~\citep{kay2017kinetics}, \emph{SSv2}~\citep{goyal2017something}, \emph{UCF101}~\citep{soomro2012ucf101}, \emph{HMDB51}~\citep{kuehne2011hmdb}, and \emph{Ego4D}~\citep{grauman2022ego4d}.  
We provide the hyperparameter setup for pre-training and fine-tuning in~\Cref{tab:hyperparams_ucf_hmdb_pt} and \Cref{tab:hyperparams_ucf_hmdb_ft}, respectively. During fine-tuning, we follow segment-based sampling~\citep{wang2018temporal} on SSv2.
As we mentioned in~\Cref{subsec:ours-tokenselect} of the main paper, we adopt the same masking ratio with VideoMAE~\citep{tong2022videomae} and ST-MAE~\citep{feichtenhofer2022masked} for masking input video, but we only reconstruct $(\rho_{pre}\times{100})\%$ of input video tokens during pre-training. And in fine-tuning, our model only takes $(\rho_{pre}\times{100})\%$ of input video tokens. We follow the linear learning rate scheduling of~\cite{tong2022videomae} and~\cite{feichtenhofer2022masked}, $lr = base\_learning\_rate \times batch\_size/256$. Overall implementation of our method is built upon VideoMAE\footnote{\url{https://github.com/MCG-NJU/VideoMAE}}\citep{tong2022videomae}. 
\paragraph{Video Action Recognition} We extensively evaluated our method on multiple benchmark datasets for the video action recognition task, including K400, SSv2, UCF101, and HMDB51. 
For inference, we adapt common multi-view testing with T clips $\times$ three crops. That is, the model takes T temporal clips with three spatial crops to cover the overall length and space of the video. Then, we measure the average performance of all views.
\paragraph{Object State Change Classification~(OSCC)} The OSCC dataset is the subset of the Ego4d dataset, consisting of 41.1k/21.2k train/val 8-second videos. Note that, as the Ego4D dataset shows the characteristic of having motion cues that are distributed and containing multi objects in~\Cref{fig:masking_vis} (c) of the main paper, we adopt $\rho_{pre}=0.5$.
We train 400 and 30 epochs with a fixed sampling ratio of 4 and 10 in the pre-training and fine-tuning phases, respectively. We use $\alpha=1.5$ only in pre-training. 
\section{Comparing Functions for Computing Token Embedding Distance \label{sec:supple:dist_function}}
\begin{table}[h]
\centering
% \vspace{-0.15in}
\caption{{Fine-tuning results on HMDB51 measured by varying the distance function in pre-training.}}
% \vspace{-0.1in}
%\addtolength{\tabcolsep}{-2.8pt}  
\renewcommand{\arraystretch}{1.0}
\renewcommand{\tabcolsep}{10pt}
\begin{tabular}{cc}

\toprule
{Distance Function} & {accuracy($\%$)}\\
\midrule
negative cosine&33.53\\
negative CKA&34.58\\
L1&42.22\\
L2&\textbf{42.81}\\
\bottomrule
% \hline \hline

\end{tabular}
\label{tab:distance_function}
% \vspace{-0.15in}
\end{table}
\begin{table*}
\centering
    \caption{\textbf{The ratio of our ReRo Masking at pre-training (Left) and fine-tuning (Right)} on UCF101.
    % $^\dagger$ adopts a half batch size to others due to the memory allocation limitation. 
    We highlight the default redundancy-robust masking rate ($\rho_{pre}$) as \highlight{red texts}. We basically pre-train our model 800 epochs, but also report the results with 3,200 pre-training epochs following VideoMAE, denoted by $^\dagger$. $^\ast$ and $^{\ast\ast}$ denotes the results with 75\% and 90\% masking, respectively.}
    \label{tab:masking_ratio}
% \vspace{-0.1in}
\begin{minipage}{0.7\columnwidth}
    % \vspace{-0.15in}
            % \vspace{0.05in}
\centering
    \resizebox{1.0\linewidth}{!}{
        \renewcommand{\arraystretch}{0.8}
        \renewcommand{\tabcolsep}{3pt}
        \begin{tabular}{c c c c}
\toprule
        % \vspace{-0.02in}
        \textbf{Method} &\textbf{$\rho_{pre}$} & \textbf{top-1} & \textbf{GFLOPs} \\
        \midrule
        % MAE \cite{feichtenhofer2022masked} &{ -} & { -} & { 46.10}\\
        MAE &{ -} & { -} & { 46.10}\\
        % \multirow{2}{*}{VideoMAE$^\dagger$ \cite{tong2022videomae}}&{ -} & {91.25}$^\ast$ & { 57.50}\\
        \multirow{2}{*}{VideoMAE$^\dagger$}&{ -} & {91.25}$^\ast$ & { 57.50}\\
        \vspace{0.03in}
        &{ -}&{90.80$^{\ast\ast}$}&{ 35.48}\\
        \midrule
        \vspace{0.03in}
        \multirow{ 4}{*}{Ours} &{ 0.5} & { 90.91} & { 23.43}\\
        \vspace{0.03in}
        &{ 0.4} & { 91.51} & { 21.54}\\
        \vspace{0.03in}
        &{ \highlight{0.3}} & \textbf{ 91.56} & { 19.81}\\
        \vspace{0.03in}
        &{ 0.2} & { 90.80} & { \textbf{18.22}}\\
        \midrule
        Ours$^\dagger$ &{ \highlight{0.3}} & \textbf{93.39} & { 19.81}\\
\bottomrule
\end{tabular}}
    % \vspace{+0.1in} 
\end{minipage}
% \hfill
\hspace{0.35in}
\begin{minipage}{0.73\columnwidth}
    % \vspace{-0.1in}
        \vspace{0.05in}
\centering
        \resizebox{1.0\textwidth}{!}{
        \renewcommand{\arraystretch}{0.8}
        \renewcommand{\tabcolsep}{3pt}
        \begin{tabular}{c c c c}
        \toprule
        \textbf{Method} &$\rho_{pre}$ & \textbf{top-1} & \textbf{GFLOPs}\\
        \midrule
        % MAE \cite{feichtenhofer2022masked} &{ -} & { -} & { 180.6}\\
        MAE &{ -} & { -} & { 180.6}\\
        % VideoMAE$^\dagger$\cite{tong2022videomae} &{ -} & { 91.25} & { 180.5}\\
        VideoMAE$^\dagger$ &{ -} & { 91.25} & { 180.5}\\
        \midrule
        \multirow{ 5}{*}{Ours} &{ 1.0} & { 89.71} & { 180.5}\\
        &{ 0.8} & { 90.17} & { 137.5}\\
        &{ 0.7} & { 91.25} & { 117.3}\\
        &{ \highlight{0.6}} & { \textbf{91.56}} & { 98.1}\\
        &{ 0.5} & { 91.48} & { \textbf{79.8}}\\
        \midrule
        Ours$^\dagger$ &{ \highlight{0.6}} & \textbf{93.39} & { 98.1}\\
\bottomrule
\end{tabular}}
\end{minipage}
% \vspace{-0.1in}
\end{table*} 

We analyze the effect of adopting different distance functions: L1 and L2 distance, negative cosine similarity, and negative CKA~\citep{cortes2012algorithms}. We train ViT-B on HMDB51 for 200 and 50 epochs, respectively, only varying the distance function in pre-training as shown in Table~\ref{tab:distance_function}. As shown, L1 and L2 outperform the alternatives. 
Therefore, we default to the L2 distance to measure the disparity between token embeddings in temporally adjacent frames.

\section{Additional Experimental Results \label{sec:supple:Additional Experimental Results}}
\paragraph{Ablation study on redundancy-robust token selection} One of our major contributions is the significantly enhanced computational efficiency during VRL, as we process a few latent vectors in the decoder to reconstruct only the motion-activated tokens in the given videos according to the redundancy-robust masking ratio $\rho_{pre}$ (Please see~\Cref{subsec:ours-tokenselect}). We set $\rho_{pre}$ to $0.3$ so that our EVEREST reconstructs the $30\%$ of the essential spatiotemporal regions focusing on objects' movements and behaviors, from a sparsified video clip, which reduces $65.5\%$ of GFLOPs at the pre-training phase (Please see~\Cref{fig:a} \highlight{Right} and \Cref{tab:masking_ratio} \highlight{Left}). Additionally, as reported in~\Cref{tab:masking_ratio} \highlight{Right}, our approach can reduce the computational overhead at the fine-tuning phase, which is practically useful when transferring the learned representation learning model to downstream video tasks.
\begin{table}[h]
\centering
\caption{\textbf{Fine-tuning performance of EVEREST on K400 using the VideoMAE pre-trained model}. We adopt our redundancy-robust token selection method for fine-tuning K400. We use the pre-trained weights by VideoMAE for 1600 epochs and Vanilla indicates a standard fine-tuning of the video action recognition task leveraging all visual tokens.}
% \vspace{-0.1in}
\resizebox{\columnwidth}{!}{\begin{tabular}{llclc}
\toprule
Pre-training  & Fine-tuning    & Backbone & GFLOPs & Acc.~(\%) \\ \midrule
\multirow{6}{*}{VideoMAE}   & Vanilla & ViT-S    & 57.0    & \textbf{79.0}       \\
& \cellcolor{gg}+\textbf{EVEREST} & \cellcolor{gg}ViT-S    & \cellcolor{gg}\textbf{29.1} \cellcolor{gg}\textcolor{blue}{($\downarrow48.9\%$)}   &\cellcolor{gg}\textbf{78.8}      \\
 & Vanilla & ViT-B    & 180.5  & \textbf{81.5}    \\
&\cellcolor{gg}+\textbf{EVEREST} & \cellcolor{gg}ViT-B    & \cellcolor{gg}\textbf{98.1} \textcolor{blue}{($\downarrow45.7\%$)}   & \cellcolor{gg}\textbf{81.6}       \\
 & Vanilla & ViT-L    & 597.2  & \textbf{85.2}    \\
&\cellcolor{gg}+\textbf{EVEREST} & \cellcolor{gg}ViT-L    & \cellcolor{gg}\textbf{330.0} \textcolor{blue}{($\downarrow44.7\%$)}  & \cellcolor{gg}\textbf{84.8}       \\ \bottomrule
\end{tabular}}\label{tab:videomae_pt_mat_ft}
% \vspace{-0.23in}
\end{table}
\paragraph{EVEREST fine-tuning using the pre-trained VideoMAE on K400.}
We also show the transferability of our redundancy-robust token selection by fine-tuning a pre-trained model with a different method, VideoMAE, on the K400 dataset. 
% To validate the generality of the proposed information-intensive token selection method, we finetune EVEREST using the pre-trained model by VideoMAE on K400. 
We borrow the pre-trained weights in the official repository\footnote{\url{https://github.com/MCG-NJU/VideoMAE/blob/main/MODEL_ZOO.md}}.
\Cref{tab:videomae_pt_mat_ft} shows the results with different ViT backbones.
% Even using public pre-trained models of VideoMAE, Our EVEREST shows competitive fine-tuning performance compared to VideoMAE, while significantly reducing computation cost. These results demonstrate that our EVEREST can seamlessly adopt backbones pretrained in other ways.
Even using publicly available pre-trained models from VideoMAE, our EVEREST show competitive fine-tuning performance compared to the baseline while significantly reducing the computational cost. These results demonstrate that EVEREST can seamlessly adopt an otherwise pre-trained backbone.

\begin{table}[h]
\centering
% \vspace{-0.15in}
% \addtolength{\tabcolsep}{5pt}  
\caption{Fine-tuning performance on UCF101 measured by a varying number of frames.}
\vspace{-0.05in}
\resizebox{\columnwidth}{!}{
\renewcommand{\arraystretch}{1.1}
\renewcommand{\tabcolsep}{6.5pt}
\begin{tabular}{cccccc}
\toprule
\multirow{2}{*}{Method} & \multicolumn{2}{c}{$\#$ of frames} & \multicolumn{2}{c}{GFLOPs} & \multirow{2}{*}{Acc.~($\%$)}\\
&PT&FT&PT&FT&\\
\midrule
\textbf{VideoMAE}&16&16&35.48&180.5&90.80\\
\cellcolor{gg}\textbf{EVEREST}&\cellcolor{gg}16&\cellcolor{gg}16&\cellcolor{gg}\textbf{19.81}&\cellcolor{gg}\textbf{98.1}&\cellcolor{gg}\textbf{93.39}\\
\cellcolor{gg}\textbf{EVEREST}&\cellcolor{gg}\textbf{24}&\cellcolor{gg}\textbf{24}&\cellcolor{gg}\textbf{30.65}&\cellcolor{gg}\textbf{159.4}&\cellcolor{gg}\textbf{94.42}\\

\bottomrule

\end{tabular}}
\label{tab:increasing_frames}
% \vspace{-0.2in}
\end{table}
\paragraph{Increasing frame lengths from reduced memory of our EVEREST} As shown in~\Cref{fig:resource} in the main paper, our EVEREST drastically saves the computational cost in pre-training and fine-tuning. In~\Cref{tab:increasing_frames}, we analyze to compensate for the reduced memory occupancy by increasing the number of input frames from 16 to 24 in pre-training and fine-tuning to observe the results in case of having similar computational costs with VideoMAE. We achieve $94.42\%$ accuracy on UCF101 while still having relatively lower GFLOPs than VideoMAE.

% Here, we show the practical results comparing the GPU memory usage and corresponding batch size of each masking-base video models in Table~\ref{tab:batchsize}. From this result, we strongly believe that our method would be the efficient implementation baseline for those who don't have many GPU resources in contrast to the previous works~\cite{feichtenhofer2022masked, tong2022videomae}.

% \input{contents/supplement_table_ablation_sampling.tex}
\begin{table}[h]
\centering
\begin{minipage}{0.32\columnwidth}
\centering
\caption{ Ablation study of on-the-fly information-intensive frame selection of EVEREST. }
% \vspace{-0.1in}
\resizebox{1.0\columnwidth}{!}{
\renewcommand{\arraystretch}{0.9}
\renewcommand{\tabcolsep}{4pt}
\begin{tabular}{c c}
        \toprule
        \vspace{-0.02in}
        $\alpha$ & \textbf{Acc. (\%)} \\
        \midrule
        { -} & { 73.17}\\
        { 1.3} & { 73.54}\\
        \highlight{ 1.5} & { \textbf{73.85}}\\
        { 1.8} & { 73.79}\\
        { 2.0} & { 73.80}\\
\bottomrule
\end{tabular}}
\label{tab:ablation_sampling}
\end{minipage}    
\hfill
\begin{minipage}{0.64\columnwidth}
\centering
\caption{ Transferability comparison between VideoMAE and EVEREST on K400 and SSv2 to UCF101 and HMDB51 with ViT-S and ViT-B.}
% \vspace{-0.1in}
%\addtolength{\tabcolsep}{-2.8pt}  
\resizebox{\columnwidth}{!}{
\renewcommand{\arraystretch}{1.7}
\renewcommand{\tabcolsep}{1.0pt}
\begin{tabular}{ccccc}
\toprule
\multirow{2}{*}{Method} & \multirow{2}{*}{backbone} & {Pre-train} & \multicolumn{2}{c}{Top1 Acc.($\%$)}\\
&&Dataset&UCF101&HMDB51\\
\midrule
{VideoMAE}&{ViT-S}&{K400}&{84.2}&{54.2}\\
\cellcolor{gg}{\textbf{EVEREST}}&\cellcolor{gg}{ViT-S}&\cellcolor{gg}{K400}&\cellcolor{gg}\textbf{89.2 \textcolor{blue}{($\uparrow5.0\%$)}}&\cellcolor{gg}\textbf{61.4 \textcolor{blue}{($\uparrow7.2\%$)}}\\
\midrule
{VideoMAE}&{ViT-B}&{SSv2}&{88.7}&{60.9}\\
\cellcolor{gg}{\textbf{EVEREST}}&\cellcolor{gg}{ViT-B}&\cellcolor{gg}{SSv2}&\cellcolor{gg}\textbf{92.2 \textcolor{blue}{($\uparrow3.5\%$)}}&\cellcolor{gg}\textbf{64.6 \textcolor{blue}{($\uparrow3.7\%$)}}\\

\bottomrule

\end{tabular}}
\label{tab:transfer}
\end{minipage}
% \vspace{-0.15in}
\end{table}

% \begin{table}[h]
% \centering
% \caption{\bf Transferability comparison between VideoMAE and EVEREST on K400 and SSv2 to UCF101 and HMDB51 with ViT-S and ViT-B.}
% \vspace{0.05in}
% %\addtolength{\tabcolsep}{-2.8pt}  
% \resizebox{0.55\textwidth}{!}{
% \renewcommand{\arraystretch}{1.1}
% \renewcommand{\tabcolsep}{4pt}
% \begin{tabular}{ccccc}
% \toprule
% \multirow{2}{*}{Method} & \multirow{2}{*}{backbone} & {Pre-train} & \multicolumn{2}{c}{Top1 Acc.($\%$)}\\
% &&Dataset&UCF101&HMDB51\\
% \midrule
% {VideoMAE}&{ViT-S}&{K400}&{84.2}&{54.2}\\
% {EVEREST}&{ViT-S}&{K400}&\textbf{89.2 \textcolor{blue}{($\uparrow5.0\%$)}}&\textbf{61.4 \textcolor{blue}{($\uparrow7.2\%$)}}\\
% \midrule
% {VideoMAE}&{ViT-B}&{SSv2}&{88.7}&{60.9}\\
% {EVEREST}&{ViT-B}&{SSv2}&\textbf{92.2 \textcolor{blue}{($\uparrow3.5\%$)}}&\textbf{64.6 \textcolor{blue}{($\uparrow3.7\%$)}}\\

% \bottomrule

% \end{tabular}}
% \label{tab:transfer}
% \vspace{-0.15in}
% \end{table}
\paragraph{The ratio of information-intensive frame selection} We conduct an ablation study for the rate of information-intensive frame sampling $\alpha$. We pre-train our model 100 epochs on the OSCC task by varying $\alpha$ from \emph{not using (-)} to \emph{2.0}. Interestingly, as shown in~\Cref{tab:ablation_sampling}, the performance with $\alpha$ outperforms baselines (-). And we default $\alpha=1.5$ for experiments, which shows the best performance than others.

\paragraph{Transferability of our EVEREST} We also measure the fine-tuning performance on UCF101 and HMDB51 using larger pre-training datasets in~\Cref{tab:transfer}. We perform fine-tuning experiments with 200 epochs of pre-trained models from K400 and SSv2 in~\Cref{tab:vs_videomae} and~\Cref{tab:ucf&hmdb}. Impressively, our method gains a substantial performance enhancement over all experiments compared with VideoMAE.

\begin{table}[h]
\centering
\vspace{-0.05in}
%\addtolength{\tabcolsep}{-2.8pt}  
\caption{\bf The impact of the number of Conv3d layers for capturing Redundancy-Robust Token Selection.}
% \vspace{-0.1in}?
\resizebox{1.0\columnwidth}{!}{
\renewcommand{\arraystretch}{1.2}
\renewcommand{\tabcolsep}{1pt}
\begin{tabular}{cccccccc}
\toprule
{\# of Conv3d} & {layer1} & {layer2} & {layer3}& {\# of}& {Memory} & \multirow{2}{*}{GFLOPs}& {FT}\\
{layers}&{kernel size}&{kernel size}&{kernel size}& {params. (M)}& {usage (GB)}&&{Acc. (\%)}\\
\midrule
1&2$\times16\times16$&N/A&N/A&\textbf{94.2 / 86.3}&\textbf{9.2} / 17.6&\textbf{19.8 / 98.1}&\textbf{68.1}\\
2&$2\times4\times4$&$1\times4\times4$&N/A&102.5 / 94.6&11.9 / \textbf{17.5}&34.6 / 112.9&66.7\\
3&$2\times4\times4$&$1\times2\times2$&$1\times2\times2$&97.8 / 89.9&11.3 / \textbf{17.5}&38.3 / 116.6&66.8\\

\bottomrule

\end{tabular}}
\label{tab:embedding}
% \vspace{-0.2in}
\end{table}
\paragraph{The design of the embedding layer} We perform an ablation study of our proposed EVEREST regarding the embedding layer design. In our default setting, we use a single Conv3d layer. And we adopt deeper neural networks for spatiotemporal embedding in this experiment. Note that we adjust the kernel size of each layer to maintain the output dimension. As shown in~\Cref{tab:embedding}, stacking more layers for input embedding achieves lower performance than the default setting (i.e., \# of Conv3D layer = 1) on the HMDB51 dataset. Even they require substantial additional training costs in terms of the number of parameters, GFLOPs, and memory usage in most cases of pre-training and fine-tuning. 
\begin{table}[h]
\centering
% \vspace{-0.2in}
\caption{\textbf{The impact of informative-token selection.} It shows better performances both in pre-training and fine-tuning when the model prioritizes selection with the most far-distance (descending) embedded tokens rather than near-distance (ascending) tokens.}
% \vspace{-0.1in}
\resizebox{\columnwidth}{!}{
\begin{tabular}{cccc}

\toprule
\multirow{ 2}{*}{Method}&\multicolumn{2}{c}{ReRo Masking} & {Fine-tuning}\\
&{Pre-train} & {Fine-tune} & {accuracy (\%)}\\

\midrule
\multirow{ 4}{*}{EVEREST}&
{ascending} & 
{ascending} & {60.93} \\
 
&{ascending} & 
{descending} & {73.49} \\
 
&{descending} & 
{ascending} & {75.60} \\

&{descending} & 
{descending} & {\textbf{91.56}} \\

% \hline \hline
\bottomrule
\end{tabular}}
\label{tab:selection-ablation}
% \vspace{-0.1in}
\end{table}
\paragraph{The importance of selecting tokens containing rich motion information} We select to learn temporally changing tokens based on the distance between token embeddings in adjacent frames, thereby, the tokens farther away from adjacent frames contain less redundant information. As shown in~\Cref{tab:selection-ablation}, when we reversely select near-distance tokens in pre-training and fine-tuning, it severely decreases the accuracy, and the performance is the worst if the model selects tokens via reverse strategies in both phases.

\begin{table}[h]
\centering
% \vspace{-0.08in}
\caption{{Comparison with running-cell masking strategy.}}
% \vspace{-0.1in}
%\addtolength{\tabcolsep}{-2.8pt}  
\renewcommand{\arraystretch}{1.0}
\renewcommand{\tabcolsep}{10pt}
\begin{tabular}{ccc}

\toprule
{Method} & {UCF101}& {HDMB51}\\
\midrule
Running-cell&91.0&61.4\\
\cellcolor{gg}\bf{EVEREST}&\cellcolor{gg}\bf{93.4}&\cellcolor{gg}\bf{68.1}\\
\bottomrule
% \hline \hline

\end{tabular}
\label{tab:running_cell}
% \vspace{-0.15in}
\end{table}
\paragraph{Comparison with running-cell masking strategy} We compare our method with running-cell masking~\cite{qing2023mar}, which is used in VideoMAE-V2~\cite{wang2023videomae}. We pre-trained with these masking strategies and performed conventional full-finetuning. As shown in~\Cref{tab:running_cell}, our method clearly outperforms running-cell masking, demonstrating our effectiveness against other selection strategies given the same architecture. We further show clear limitations in running-cell masking that VideoMAE-V2 suffers from visual redundancy and cannot capture semantic importance (Please see~\Cref{fig:masking-comparison}). 
 
Next, We emphasize that redundancy-robust masking is applicable not only in the pre-training phase, but also in performing video-based downstream tasks. We process only 60\% of tokens ($\rho_{pre}=0.6$) of the given video during fine-tuning. Surprisingly, our redundancy-robust masking for the downstream tasks gains increased performance than our variant without masking on the fine-tuning tasks ($\rho_{pre}=1.0$) using only about 55\% of GFLOPs, as shown in~\Cref{tab:masking_ratio} \highlight{Right}. The results support our hypothesis that video data often contain redundant information and our selective video learning using the proposed masking strategy successfully captures essential space-time regions in video inputs, allowing us to focus more on learning spatiotemporally meaningful features.
The fine-tuning masking ratio is higher than the ratio at the pre-training stage, showing that the model uses more information to fully exploit the task-relevant cues from the given videos than pre-training, which aims to obtain general information.

\section{Visualization of Redundancy-Robust Masking and Sampling \label{sec:supple:Visualization of motion-centric masking and sampling}}
\begin{figure*}[h]
    \centering
        \includegraphics[width=0.7\linewidth]{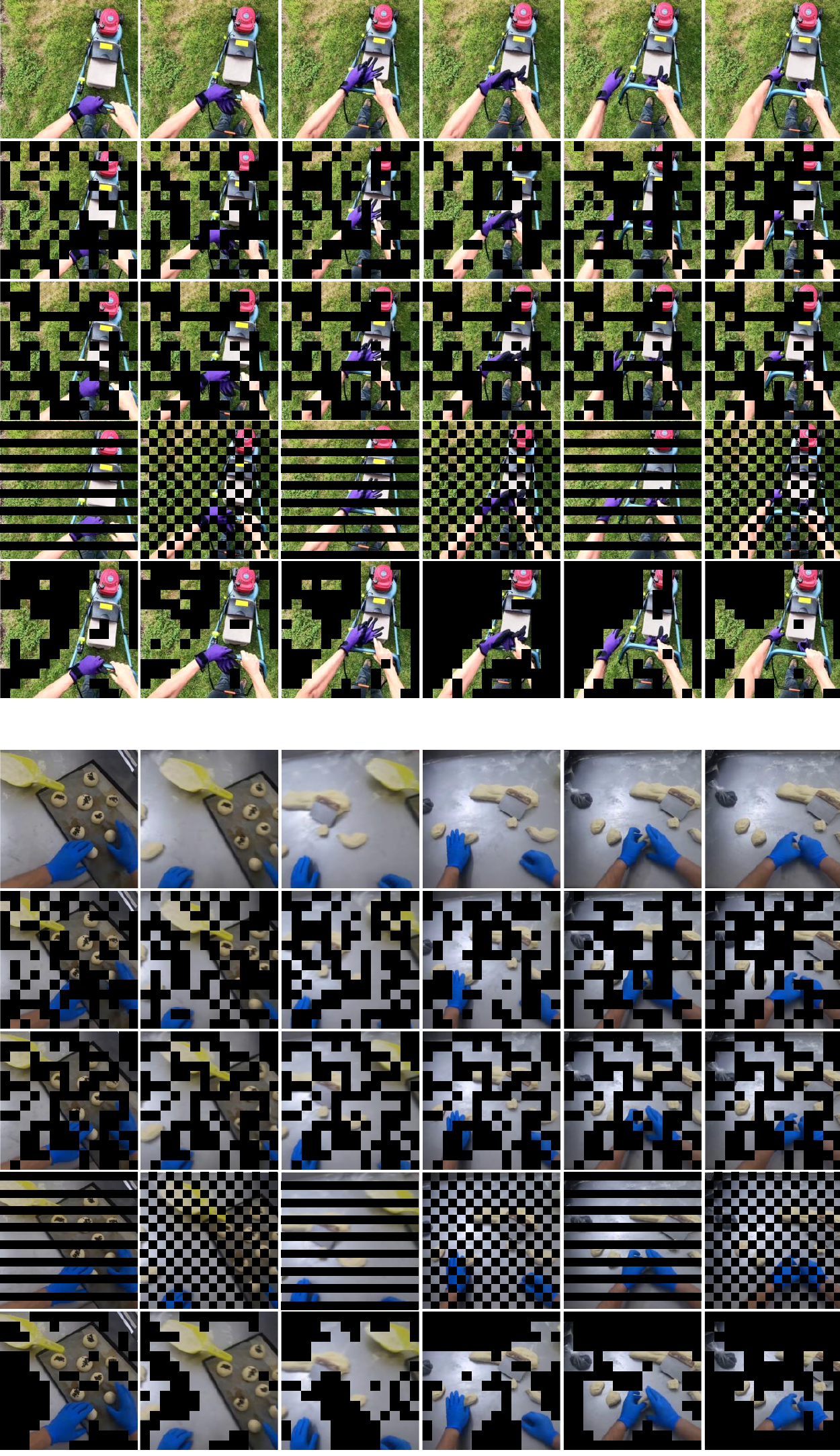}
% \vspace{-0.05in}
     \caption{{\textbf{Examples of masking strategies.} We visualize the masking strategy of ST-MAE~\citep{feichtenhofer2022masked} (random), VideoMAE~\citep{tong2022videomae} (space-only), {VideoMAE V2~\citep{wang2023videomae} (running cell~\citep{qing2023mar})}, and our EVEREST on Ego4D dataset. From the top, each row indicates the original frames, MAE masking, VideoMAE masking, {VideoMAE V2 masking}, and our proposed method, respectively. We set $\rho_{pre}$ to 0.5.}}
    \label{fig:masking-comparison}
% \vspace{0.15in}
\end{figure*}

\begin{figure*}[h]
    \centering
        \includegraphics[width=0.7\linewidth]{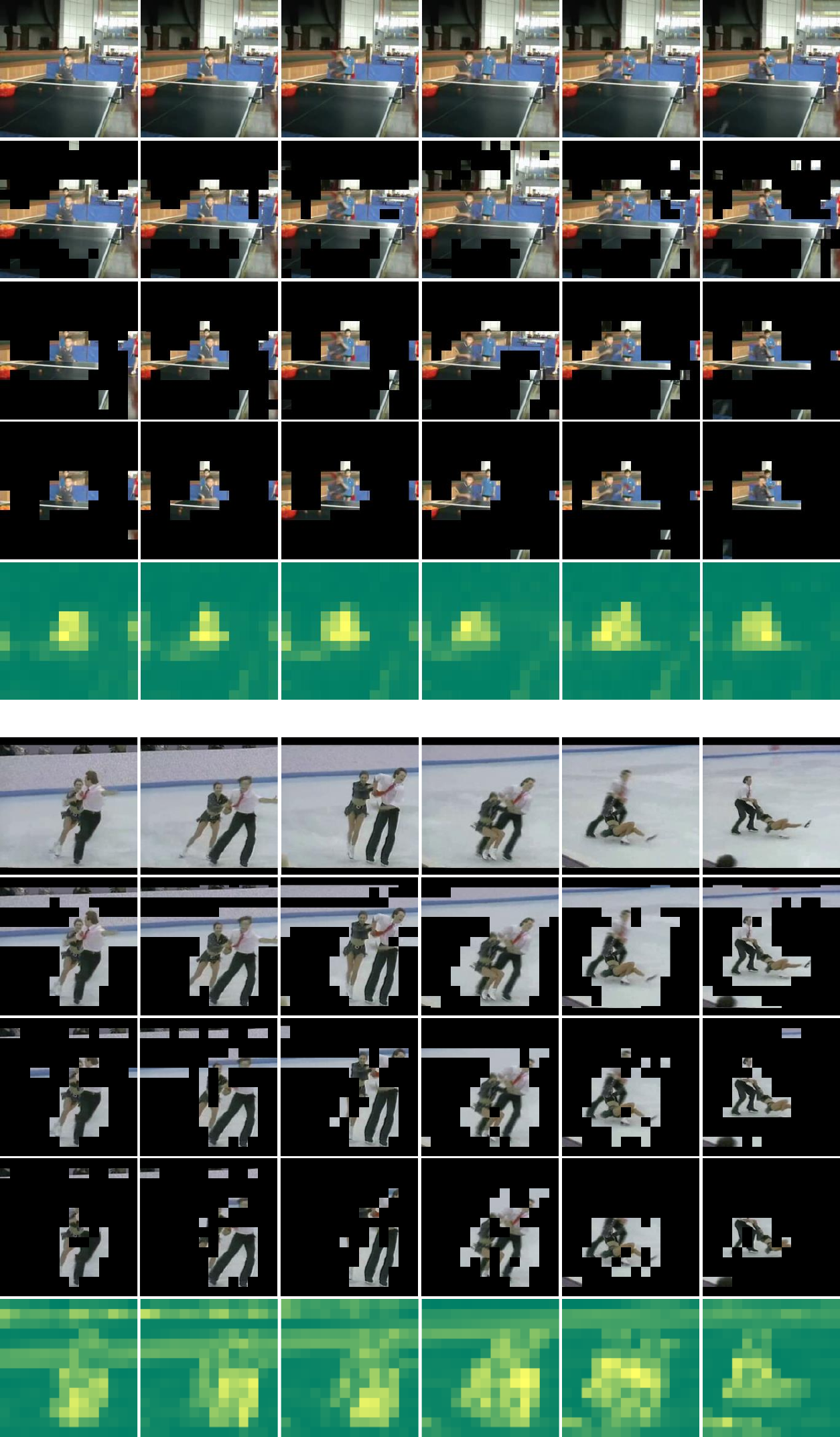}
% \vspace{-0.05in}
     \caption{{\textbf{More examples of Redundancy-robust token selection.} We show the original video frames in the first row, Redundancy-robust masking results in the second, third, and fourth row by varying the $\rho_{pre}$ to 0.5, 0.25, and 0.15, respectively, and obtained importance heatmaps in the last row on UCF101~\citep{soomro2012ucf101} dataset.}}
    \label{fig:supple_visss}
% \vspace{0.15in}
\end{figure*}

\begin{figure*}
    \centering
        \includegraphics[width=0.7\linewidth]{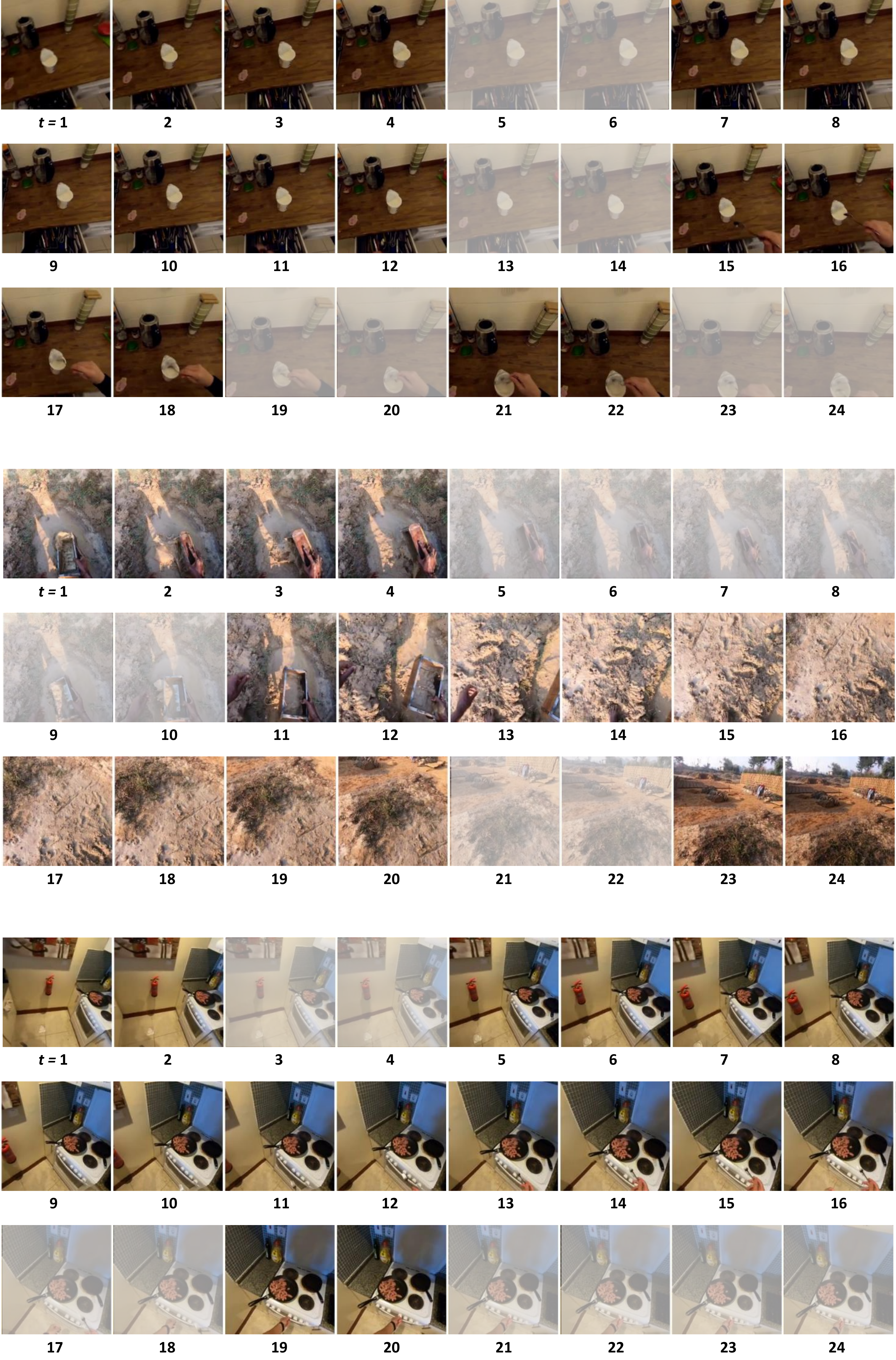}
% \vspace{0.03in}
     \caption{\textbf{Examples of on-the-fly information-intensive frame sampling based on our EVEREST.} Among 24 frames in the given video clip, our EVEREST adaptively selects frames containing rich motion information based on temporal correlation. Non-blurred frames are sampled through our method and used in the pre-training phase.}
    \label{fig:supple_vis_mcs}
\end{figure*}

As shown in~\Cref{fig:masking_vis} of the main paper, our masking method successfully captures the most informative space of each frame in various conditions. In \Cref{fig:masking-comparison}, we visualize masking strategies of other strong baselines, ST-MAE~\citep{feichtenhofer2022masked} and VideoMAE~\citep{tong2022videomae}. 
While these two methods sample masked tokens based on random probability without taking into account semantic motion information, our proposed method successfully captures temporally significant localized regions in the video. We further visualize our masking results in~\Cref{fig:supple_visss} for deeper understanding. The upper sample is similar in~\Cref{fig:masking_vis} (b) but changes the $\rho_{pre}$ from 0.5 to 0.15. The masked result shows that our masking strategy could be better and save more computational costs if the view is fixed and the moving object is few. The bottom sample show that our masking strategy wrongly captured the blue line of each frame as informative tokens when $\rho_{pre}$ is relatively large(=0.5). However, when $\rho_{pre}$ is relatively small(=0.25, 0.15), it ignores the blue line of each frame and concentrates more on moving objects in the given video. In \Cref{fig:supple_vis_mcs}, We show selected frames via our adaptive frame sampling. 
Our method uses redundancy-robust token selection to draw frames to learn in an online manner based on the proportion that contains important tokens containing fluent motion information or significant change of temporal states. This strategy allows the model to take diverse frames from given videos while discarding redundant or low-information frames.


%%%%%%%%%%%%%%%%%%%%%%%%%%%%%%%%%%%%%%%%%%%%%%%%%%%%%%%%%%%%
\end{document}